\documentclass[journal, 12pt]{IEEEtran}

\usepackage{etoolbox}
\usepackage{capt-of}
\makeatletter
\let\NAT@parse\undefined
\apptocmd\@maketitle{{\eyecatcher{}\par}}{}{}
\makeatother

\usepackage{amsmath}
\usepackage{amsfonts}
\usepackage{graphicx}
\usepackage{siunitx}
\usepackage{xcolor}
\usepackage{multirow}
\usepackage{booktabs}
\usepackage{nameref}
\usepackage[numbers, square, comma,compress]{natbib}
\usepackage[skip=2pt,font=small]{caption}
\usepackage{balance}
\usepackage{hyperref}

\definecolor{somegray}{rgb}{0.5, 0.5, 0.5}
\newcommand{\darkgrayed}[1]{\textcolor{somegray}{#1}}
\makeatletter
\newcommand*\titleheader[1]{\gdef\@titleheader{#1}}
\AtBeginDocument{%
  \let\st@red@title\@title
  \def\@title{%
    \vskip-2.0em
    \bgroup\normalfont\large\centering\@titleheader\par\egroup
    \vskip0.0em\st@red@title}
}

\makeatother

\titleheader{\darkgrayed{This paper has been accepted for publication in Science Robotics\copyright, July 21, 2021.\\
\url{https://robotics.sciencemag.org/content/6/56/eabh1221}}}

\IEEEoverridecommandlockouts

\title{\huge Time-Optimal Planning for Quadrotor Waypoint Flight} 

\author{Philipp Foehn, Angel Romero, Davide Scaramuzza
\thanks{The authors are with the Robotics and Perception Group, University of Zurich, Switzerland (\protect\url{http://rpg.ifi.uzh.ch}).
This work was supported by the National Centre of Competence in Research (NCCR) Robotics through the Swiss National Science Foundation (SNSF) and the European Union’s Horizon 2020 Research and Innovation Programme under grant agreement No. 871479 (AERIAL-CORE) and the European Research Council (ERC) under grant agreement No. 864042 (AGILEFLIGHT).
}}

\begin{document}

\newcommand\eyecatcher{
\centering
\vspace{8pt}
\captionsetup{type=figure}\setcounter{figure}{0}
\includegraphics[width=\linewidth]{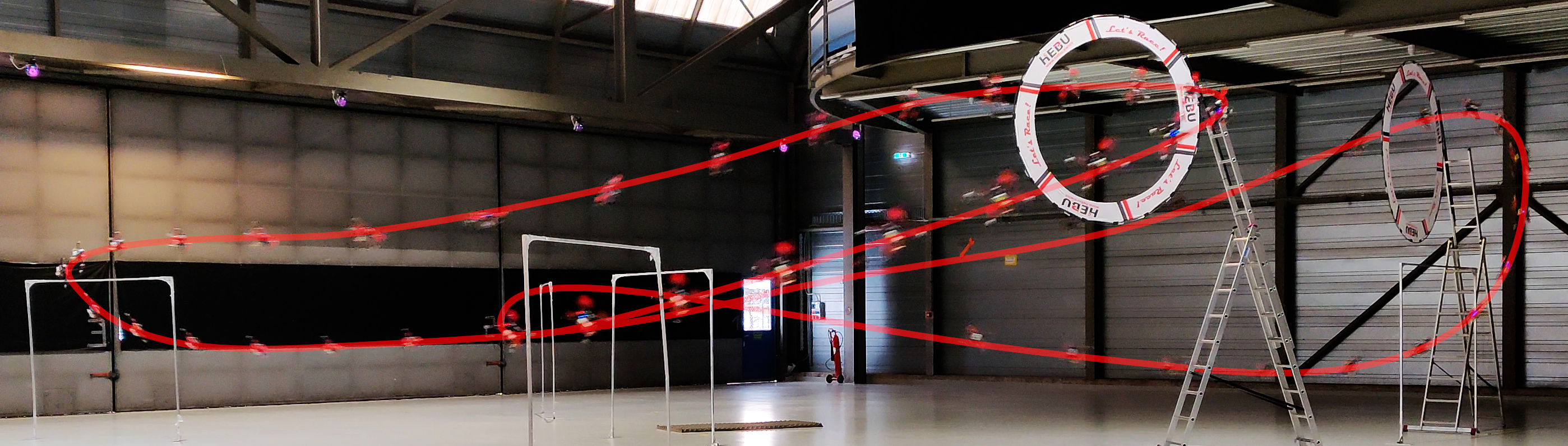}
\caption{\textbf{A time-optimal trajectory.} This time-optimal flight path was computed using the proposed complementary progress constraints (CPC) and executed in a motion capture system, outperforming the best human expert.}
\label{fig:eyecatcher}
\vspace{-20pt}
}

\maketitle 

\begin{abstract}
Quadrotors are amongst the most agile flying robots.
However, planning time-optimal trajectories at the actuation limit through multiple waypoints remains an open problem.
This is crucial for applications such as inspection, delivery, search and rescue, and drone racing.
Early works used polynomial trajectory formulations, which do not exploit the full actuator potential due to their inherent smoothness. 
Recent works resorted to numerical optimization, but require waypoints to be allocated as costs or constraints at specific discrete times.
However, this time-allocation is a priori unknown and renders previous works incapable of producing truly time-optimal trajectories.
To generate truly time-optimal trajectories, we propose a solution to the time allocation problem while exploiting the full quadrotor's actuator potential.
We achieve this by introducing a formulation of progress along the trajectory, which enables the simultaneous optimization of the time-allocation and the trajectory itself.
We compare our method against related approaches and validate it in real-world flights in one of the world's largest motion-capture systems, where we outperform human expert drone pilots in a drone-racing task.
\end{abstract}

\section*{Multimedia Material}
A video demonstrating the flight experiments can be found at \url{https://youtu.be/ZPI8U1uSJUs}.
\newline
Code: \url{https://github.com/uzh-rpg/rpg_time_optimal}

\section{Introduction}
Autonomous drones are nowadays used for inspection, delivery, cinematography, search-and-rescue, and entertainment such as drone racing \cite{Loianno2020jfr}.
The most prominent aerial system is the quadrotor, thanks to its simplicity and versatility, ranging from smooth maneuvers to extremely aggressive trajectories.
This renders quadrotors amongst the most agile and maneuverable aerial robots\cite{Ackerman2020online,Verbeke18jmav}.

However, quadrotors have limited flight range, dictated by their battery capacity, which limits how much time can be spent on a specific task.
If the task consists of visiting multiple waypoints (delivery, inspection, drone racing \cite{Moon19jirc,Foehn20rss,Loquercio2019tro}), doing so in minimal time is often desired, and, in the context of search and rescue or drone racing (Fig. \ref{fig:eyecatcher}), even the ultimate goal.
In fact, expert human drone racing pilots accomplish this with astonishing performance, guiding their quadrotors through race tracks at speeds so far unreached by any autonomous system.
This begs the question of how close human pilots fly to the theoretical limit of a quadrotor, and whether planning algorithms could find and execute such theoretical optima.

For simple point-mass systems, time-optimal trajectories can be computed in closed-form, resulting in bang-bang acceleration trajectories \cite{LaValle2006book}, which can be sampled over multiple waypoints \cite{Pontryagin1962wiley}.
However, quadrotors are underactuated systems that need to rotate to adjust their actuated acceleration direction, which always lies in the body $z$-axis \cite{Bouabdallah07iros,Mahony12ram}.
Both the linear and rotational acceleration are controlled through the rotor thrusts, which are physically limited by the actuators.
This introduces a coupling in the achievable linear and rotational accelerations.
Therefore, time-optimal planning becomes the search for the optimal tradeoff between maximizing these accelerations.

\subsection{Related Work}
Two common approaches for planning quadrotor trajectories exist, continuous-time polynomials and discrete-time state space representations.
The first option is the widely used polynomial formulation \cite{Mahony12ram,Mellinger12ijrr,Mueller13iros} exploiting the quadrotor's differentially-flat output states with high computational efficiency.
However, these polynomials are inherently smooth and therefore cannot represent rapid state or input changes (e.g. bang-bang \cite{LaValle2006book}) at reasonable order, and only reach the input limits for infinitesimal short durations, or constantly for the full trajectory time.
This renders polynomials suboptimal since they cannot exploit the full actuator potential.
Both problems are visualized and further explained in Section \ref{sec:preface}, in the supplementary material.

The second option includes all approaches using time-discretized trajectories which can be found using search and sampling-based methods \cite{Webb13icra,Allen16gnc,Liu2018ral,Zhou19ral} or optimization-based methods \cite{Hargraves87jgcd,Geisert16icra,Augugliaro2012iros,Penin17iros}.
However, sampling the $4$-dimensional continuous input space over many discrete time steps with sufficient resolution quickly becomes computationally intractable, which is why prior work \cite{Webb13icra,Allen16gnc,Liu2018ral,Zhou19ral} restores to point-mass, polynomial, or differential-flatness approximations, and therefore does not handle single-rotor thrust constraints.
Therefore, planning time-discretized trajectories with optimization-based methods is the only viable solution in the short-medium term.
In such methods, the system dynamics and input boundaries are enforced as constraints.
In contrast to the polynomial formulation, this allows the optimization to pick any input within bounds for each discrete time step.
For a time-optimal solution, the trajectory time $t_N$ is part of the optimization variables and is the sole term in the cost function.
However, if multiple waypoints must be passed, these must be allocated as constraints to specific nodes on the trajectory.
This time allocation is a priori undefined, since the time spent between any two waypoints is unknown, which renders traditional discretized state space formulations ineffective for time-optimal trajectory generation through multiple waypoints.
    
We investigate this problem and provide a solution that allows simultaneously optimizing the trajectory and waypoint allocation in a given sequence, exploiting the full actuator potential of a quadrotor.
Our approach formulates a progress measure for each waypoint along the trajectory, indicating completion of a waypoint (see Fig. \ref{fig:cpc}).
We then introduce a "Complementary Progress Constraint" (CPC), that allows completion only in proximity to a waypoint.
Intuitively, proximity and progress must complement each other, enforcing completion of all waypoints without specifying their time allocation.

There already exists a number of works towards time-optimal quadrotor flight \cite{Hehn12ar,Loock13ecc,Spedicato18tcst,Ryou2020arxiv}, which, however, all suffer from severe limitations, such as limiting the collective thrust and bodyrates, rather than the actual constraint of limited single rotor thrusts.

The two earlier works \cite{Hehn12ar,Loock13ecc} are based on the aforementioned bang-bang approaches extended through numerical optimization of the switching times \cite{Hehn12ar} and a trajectory representation using a convex combination of multiple analytical path functions \cite{Loock13ecc}.
However, both are restricted to 2-dimensional maneuvers, whereas our approach generalizes to arbitrary 3D waypoint sequences.

Another approach is taken in \cite{Spedicato18tcst}, where a change of variables along an analytic reference path is used to put the vehicle state space into a traverse-dynamics formulation.
This allows using the arc length along the reference path as a progress measure and enables the formulation of costs and constraints independent of the time variable.
However, as in the previous works, they simplify the platform limits to collective thrust and bodyrates, neglecting realistic actuator saturation.
Furthermore, due to the use of Euler angles, their orientation space only covers a subset of the feasible attitudes and limits the solutions to a, possibly sub-optimal, subspace.

Finally, \cite{Ryou2020arxiv} uses a completely different approach, where the segment times of a polynomial trajectory are refined based on learning a Gaussian Classification model predicting feasibility.
The classification is trained on analytic models, simulation, and real flight data.
While emphasizing real-world applicability, this approach is still constrained to polynomials and requires real-world data specifically collected for the given vehicle.
Furthermore, it is an approximate method which only refines the execution speed of a predefined trajectory, rather than modifying the trajectory itself to a time-optimal solution, as opposed to our method.

\subsection{Contribution}
\begin{figure}[t]
    \centering
    \includegraphics[width=\linewidth]{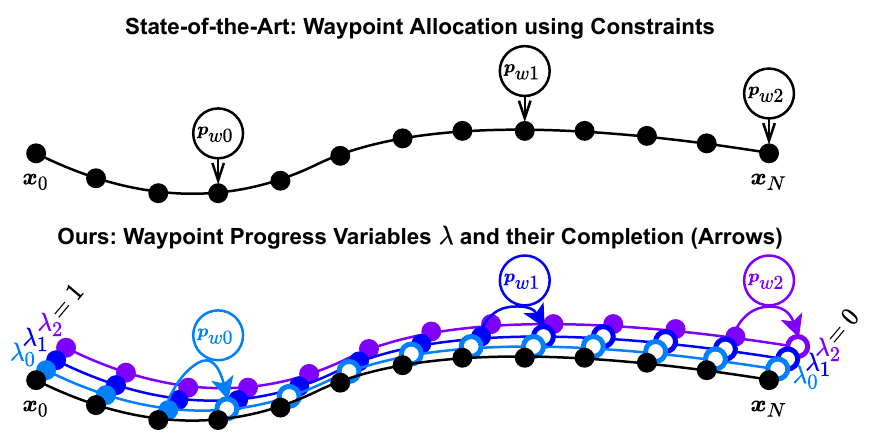}
    \caption{\textbf{Progress variables.} Top: state-of-the-art fixed allocation of positional waypoints $\mathbf{p}_{wj}$ to specific nodes $\mathbf{x}_i$.
    Bottom: our method of defining one progress variable $\lambda_j$ per waypoint.
    The progress variable can switch from $1$ (incomplete) to $0$ (completed) only when in the proximity of the relevant waypoint, implemented as a complementary constraint.}
    \label{fig:cpc}
\end{figure}

In contrast to existing methods, our approach resolves these problems by taking inspiration from optimization under contacts \cite{Posa2014ijrr}, proposing the formulation of \textit{complementary progress constraints}, where we introduce a measure of progress and complement \cite{Cottle1968laa} it with waypoint proximity.
More specifically, we formulate two factors that must complement each other, where, in our case, one factor is the completion of a waypoint (progress), while the other factor is the local proximity to a waypoint (Fig. \ref{fig:cpc}).
Intuitively, a waypoint can only be marked as completed when the quadrotor is within a certain tolerance of the waypoint (Fig. \ref{fig:complementary}), allowing simultaneous optimization of the state and input trajectory, and the waypoint time allocation.

We demonstrate how our formulation can generate trajectories that are faster than human expert flights and evaluate it against two professional human drone racing pilots, outperforming them in terms of lap time and consistency on a 3D race track in a large-scale motion capture system (Fig. \ref{fig:eyecatcher}).
Since our proposed optimization problem is highly non-convex, we also provoke non-convexity effects in simulation experiments in the supplementary material (Section \ref{sec:experiment_trajgen}).

Our method can not only serve as a baseline for time-optimal quadrotor flight but might also find applications in other fields, such as (multi-) target interception, orbital maneuvers, avoiding mixed-integer formulations \cite{Richards2002acc}, and any problem where a sequence of task goals of unknown duration must be optimized under complex dynamic constraints.
\section{Methodology}
\label{sec:method}
\subsection*{General Trajectory Optimization}
The general optimization problem of finding the minimizer $\mathbf{x}^*$ for cost $L(\mathbf{x})$ in the state space $\mathbf{x} \in \mathbb{R}^n$ can be stated as
\begin{align}
\mathbf{x}^* = \underset{\mathbf{x}}{\text{arg~min}}&~L(\mathbf{x}) \\
\text{subject to} \quad
\mathbf{g}(\mathbf{x}) &= 0 
\quad \text{and} \quad
\mathbf{h}(\mathbf{x}) \leq 0 \nonumber
\end{align}
where $\mathbf{g}(\mathbf{x})$ and $\mathbf{h}(\mathbf{x})$ contain all equality and inequality constraints respectively.
The full state space $\mathbf{x}$ is used equivalently to the term \textit{optimization variables}.
The cost $L(\mathbf{x})$ typically contains one or multiple quadratic costs on the deviation from a reference, costs on the systems actuation inputs, or other costs describing any desired behaviors.

\subsubsection{Multiple Shooting Method}
To represent a dynamic system in the state space, the system state $\mathbf{x}_k$ is described at discrete times $t_k = dt \cdot k$ at $k \in [0, N]$, also called nodes, where its actuation inputs between two nodes are $\mathbf{u}_k$ at $t_k$ with $k \in [0, N)$.
The systems evolution is defined by the dynamics $\dot{\mathbf{x}} = \mathbf{f}_{dyn}(\mathbf{x}, \mathbf{u})$,
anchored at $\mathbf{x}_0=\mathbf{x}_{init}$,
and implemented as an equality constraint of the 4th-order Runge-Kutta integration scheme (\textit{RK4}):
\begin{equation}
\mathbf{x}_{k+1} - \mathbf{x}_k - dt \cdot \mathbf{f}_{RK4}(\mathbf{x}_k, \mathbf{u}_k) = 0
\label{eqn:ms_forwardeuler}
\end{equation}
which is part of $\mathbf{g}(\mathbf{x})=0$ in the general formulation.
Both $\mathbf{x}_k$, $\mathbf{u}_k$ are part of the state space and can be summarized as the vehicle's dynamic states $\mathbf{x}_{dyn,k}$ at node $k$.
Note that this renders the problem formulation non-convex for non-linear system dynamics.

\subsection*{Time-Optimal Trajectory Optimization}
Optimizing for a time-optimal trajectory means that the only cost term is the overall trajectory time $L(\mathbf{x})=t_N$.
Therefore, $t_N$ needs to be in the optimization variables $\mathbf{x} = [t_N, \dots]^\top$, and must be positive $t_N > 0$.
The integration scheme can then be adapted to use $dt = t_N / N$.

\subsection{Passing Waypoints through Optimization}
To generate trajectories passing through a sequence of waypoints $\mathbf{p}_{wj}$ with $j \in [0, \dots, M]$, one would typically define a distance cost or constraint and allocate it to a specific state $\mathbf{x}_{dyn, k}$ at node $k$ with time $t_k$.
For cost-based formulations, quadratic distance costs are robust in terms of convergence and implemented as
\begin{equation}
L_{dist, j} = (\mathbf{p}_k - \mathbf{p}_{wj})^\top (\mathbf{p}_k - \mathbf{p}_{wj})    
\end{equation}
where $\mathbf{p}_k$, part of $\mathbf{x}$, is the position state at a user defined time $t_k$.
However, such a cost-based formulation is only a soft requirement and if summed with other cost terms does not imply that the waypoint is actually passed within a certain tolerance.
To guarantee to pass within a tolerance, constraint-based formulations can be used, such as
\begin{equation}
(\mathbf{p}_k - \mathbf{p}_{wj})^\top (\mathbf{p}_k - \mathbf{p}_{wj}) \leq \tau_j^2    
\end{equation}
which in the general problem is part of $\mathbf{h}(\mathbf{x}) \leq 0$, and requires the trajectory to pass by waypoint $j$ at position $\mathbf{p}_{wj}$ within tolerance $\tau_j$ at time $t_k$.

\subsection{Progress Measure Variables}
To describe the progress throughout a track we want a measure that fulfills the following requirements: (I) it starts at a defined value, (II) it must reach a different value by the end of the trajectory, and (III) it can only change when a waypoint is passed within a certain tolerance.
To achieve this, let the vector $\boldsymbol{\lambda}_k \in \mathbb{R}^M$ define the progress variables $\lambda_k^j$ at timestep $t_k$ for all $M$ waypoints indexed by $j$.
All progress variables start at 1 as in $\boldsymbol{\lambda}_0 = \mathbf{1}$ and must reach 0 at the end of the trajectory as in $\boldsymbol{\lambda}_N = \mathbf{0}$.
The progress variables $\boldsymbol{\lambda}$ are chained together and their evolution is defined by
\begin{equation}
\boldsymbol{\lambda}_{k+1} = \boldsymbol{\lambda}_k - \boldsymbol{\mu}_k
\end{equation}
where the vector $\boldsymbol{\mu}_k \in \mathbb{R}^M$ indicates the progress change at every timestep.
Note that the progress can only be positive, therefore $\mu_k^j \geq 0$.
Both $\boldsymbol{\lambda}_k$ and $\boldsymbol{\mu}_k$ for every timestep are part of the optimization variables $\mathbf{x}$, which replicates the multiple shooting scheme for the progress variables.
To define when and how the progress variables can change, we now imply a vector of constraints $\mathbf{f}_{prog}$ on $\boldsymbol{\mu}_k$, in its general form as
\begin{align}
\boldsymbol{\epsilon}^-_k \leq \mathbf{f}_{prog}(\mathbf{x}_k, \boldsymbol{\mu}_k) \leq \boldsymbol{\epsilon}^+_k
\end{align}
where $\boldsymbol{\epsilon}^-$, $\boldsymbol{\epsilon}^+$ can form equality or inequality constraints.
Finally, to ensure that the waypoints are passed in the given sequence, we enforce subsequent progress variables to be bigger than their prequel at each timestep by
\begin{equation}
\lambda_k^j \leq \lambda_k^{j+1} \quad \forall \quad k \in [0, N], j \in [0, M).
\end{equation}

Note that the last waypoint $\mathbf{p}_{wM}$ is always reached at the last node at $t_N$, and, therefore, could be implemented as a fixed positional constraint on $\mathbf{x}_N$, without loss of generality.

\subsection*{Complementary Progress Constraints}
In the context of waypoint following, the goal is to allow $\boldsymbol{\mu}_k$ to only be non-zero at the time of passing a waypoint.
Therefore, $\mathbf{f}_{prog}$ and $\boldsymbol{\epsilon}^- = \boldsymbol{\epsilon}^+ = 0$ are chosen to represent a \textit{complementarity constraint} \cite{Posa2014ijrr}, as
\begin{align}
f_{prog,j}(\mathbf{x}_k, \boldsymbol{\mu}_k) &= \mu_k^j \cdot \|\mathbf{p}_k - \mathbf{p}_{wj}\|_2^2 := 0 \nonumber \\
\forall j &\in [0, M]
\label{eqn:compconst}
\end{align}
which can be interpreted as a mathematical \textit{NAND (not and)} function, since either $\mu_k^j$ or $\|\mathbf{p}_k - \mathbf{p}_{wj}\|$ must be $0$.
Intuitively, the two elements \textit{complement} each other.

\begin{figure}[t]
    \centering
    \includegraphics[width=\linewidth]{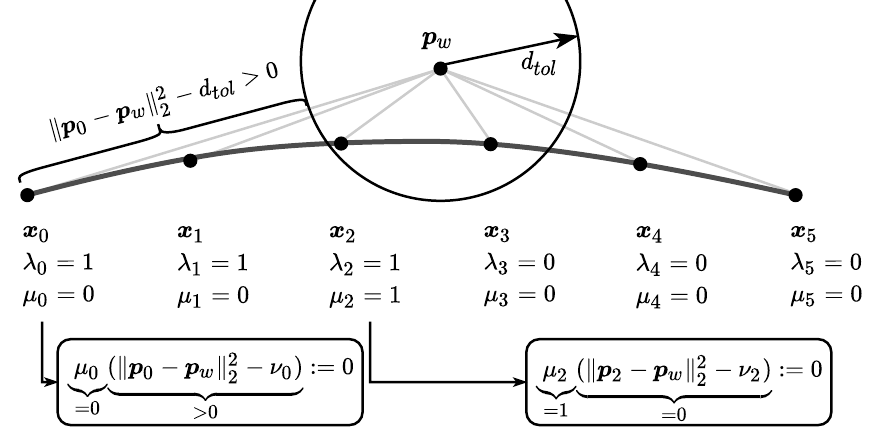}
    \caption{\textbf{Complementary progress constraint.}
    The progress change $\mu$ can only be non-zero if the distance to the waypoint $\mathbf{p}_w$ is less than the tolerance $d_{tol}$.
    This is not the case for $\mathbf{x}_0$, but for $\mathbf{x}_1$, and allowing the progress variable to switch to $0$ (complete).}
    \label{fig:complementary}
\end{figure}

\subsection{Tolerance Relaxation}
\label{sec:cpc_tol}
With \eqref{eqn:compconst} the trajectory is forced to pass \textit{exactly} through a waypoint.
Not only is this impractical, since often a certain tolerance is admitted or even wanted, but it also negatively impacts the convergence behavior and time-optimality, since the system dynamics are discretized and one of the discrete timesteps must coincide with the waypoint.
Therefore, it is desirable to relax a waypoint constraint by a certain tolerance which is achieved by extending \eqref{eqn:compconst} to
\begin{align}
f_{prog,j}(\mathbf{x}_k, \boldsymbol{\mu}_k) &= \mu_k^j \cdot \left( \|\mathbf{p}_k - \mathbf{p}_{wj}\|_2^2 - \nu_k^j\right) := 0 \nonumber \\
\text{subject to} \quad 0 &\leq \nu_k^j \leq d_{tol}^2
\forall j \in [0, M] \label{eqn:compconsttol}
\end{align}
where $\nu_k^j$ is a slack variable to allow the distance to the waypoint to be relaxed to zero when it is smaller than $d_{tol}$, the maximum distance tolerance.
This now enforces that the progress variables cannot change, except for the timesteps at which the system is within tolerance to the waypoint.
Furthermore, please note that the spatial discretization $\delta s$ depends on the number of nodes $N$ and the speed profile.
It should hold that $\delta s < d_{tol}$, to always allow at least one node to lie within the tolerance, as visualized in Fig. \ref{fig:complementary}.
This can be verified after the optimization and approximated beforehand by $\delta s \approx D / N$, where $D$ is the cumulative distance between all waypoints.

%%%%%%%%%%%%%%%%%%%%%%%%%%%%%%%%%%%%%%%%%%%%%%%%%%%%%%%%%%%%%%%%%%%%%%%%%%%%%%%%

\subsection{Optimization Problem Summary}
The full space of optimization variables $\mathbf{x}$ consists of the overall time and all variables assigned to nodes $k$ as $\mathbf{x}_k$.
All nodes $k$ include the robot's dynamic state $\mathbf{x}_{dyn,k}$, its inputs $\mathbf{u}_k$, and all progress variables, 
$\mathbf{x} = [t_N, \mathbf{x}_0, \dots, \mathbf{x}_N ]$ where
\begin{equation}
\mathbf{x}_k = 
\begin{cases}
[\mathbf{x}_{dyn, k}, \mathbf{u}_k, \boldsymbol{\lambda}_k, \boldsymbol{\mu}_k, \boldsymbol{\nu}_k] &  \text{for } k \in [0, N) \\
[\mathbf{x}_{dyn, N}, \boldsymbol{\lambda}_N] & \text{for } k = N.
\end{cases}
\nonumber
\end{equation}
\\
Based on this representation, we write the full problem as
\begin{gather}
\mathbf{x}^* = \underset{\mathbf{x}}{\text{arg~min}}~t_N
\end{gather}
subject to the system dynamics and initial constraint
\begin{align}
\mathbf{x}_{k+1} - \mathbf{x}_k - dt \cdot \mathbf{f}_{RK4}(\mathbf{x}_k, \mathbf{u}_k) &= 0 &
\mathbf{x}_0 &= \mathbf{x}_{init},
\nonumber
\end{align}
the input constraints
\begin{align}
\mathbf{u}_{min} - \mathbf{u}_k &\leq 0 &
\mathbf{u}_k - \mathbf{u}_{max} &\leq 0,
\end{align}
the progress evolution, boundary, and sequence constraints 
\begin{gather}
\boldsymbol{\lambda}_{k+1} - \boldsymbol{\lambda}_k + \boldsymbol{\mu}_k = \mathbf{0} \nonumber \\
\begin{aligned}
\boldsymbol{\lambda}_0 -1 &= \mathbf{0} & \quad
\boldsymbol{\lambda}_N &= \mathbf{0}
\end{aligned} \\
\begin{aligned}
\boldsymbol{\mu}_k & \geq 0 &
\lambda_k^j - \lambda_k^{j+1} &\leq 0 &
\forall k &\in [0, N), & j &\in [0, M), \nonumber
\end{aligned}
\end{gather}
and the complementary progress constraint with tolerance
\begin{gather}
\mu_k^j \cdot \left( \| \mathbf{p}_k - \mathbf{p}_{wj} \|_2^2 - \nu_k^j \right) = 0 \label{eq:cpc}\\
\begin{aligned}
-\nu_k^j &\leq 0 & \quad
\nu_k^j - d_{tol}^2 &\leq 0.
\label{eq:cpc_slack}
\end{aligned}
\end{gather}
Note that constraints (\ref{eq:cpc}, \ref{eq:cpc_slack}) are non-linear due to the norm on distance and the bilinearity of the progress change $\boldsymbol{\mu}$ and tolerance slack $\boldsymbol{\nu}$.

%%%%%%%%%%%%%%%%%%%%%%%%%%%%%%%%%%%%%%%%%%%%%%%%%%%%%%%%%%%%%%%%%%%%%%%%%%%%%%%%

\subsection{Quadrotor Dynamics}
The quadrotor's state space is described between the inertial frame $I$ and body frame $B$, as $\mathbf{x} = [\mathbf{p}_{IB}, \mathbf{q}_{IB}, \mathbf{v}_{IB}, \boldsymbol{\omega}_{B}]^\top$ corresponding to position $\mathbf{p}_{IB}\in \mathbb{R}^3$, unit quaternion rotation on the rotation group $\mathbf{q}_{IB} \in \mathbb{SO}(3)$ given $\| \mathbf{q}_{IB} \|=1$, velocity $\mathbf{v}_{IB} \in \mathbb{R}^3$, and bodyrate $\boldsymbol{\omega}_{B} \in \mathbb{R}^3$.
The input modality is on the level of collective thrust $\mathbf{T}_B = \begin{bmatrix}0 & 0 & T_{Bz}\end{bmatrix}^\top$ and body torque $\boldsymbol{\tau}_B$.
From here on we drop the frame indices since they are consistent throughout the description.
The dynamic equations are
\begin{gather}
\begin{aligned}
\dot{\mathbf{p}} &= \mathbf{v} &
\dot{\mathbf{q}} &= \frac{1}{2} \boldsymbol{\Lambda} (\mathbf{q}) \begin{bmatrix}0 \\ \boldsymbol{\omega}\end{bmatrix} \\ 
\dot{\mathbf{v}} &= \mathbf{g} + \frac{1}{m} \mathbf{R}(\mathbf{q}) \mathbf{T} &
\quad \dot{\boldsymbol{\omega}} &= \mathbf{J}^{-1} \left( \boldsymbol{\tau} - \boldsymbol{\omega} \times \mathbf{J} \boldsymbol{\omega} \right)
\label{eq:quad_dynamics}
\end{aligned}
\end{gather}
where $\boldsymbol{\Lambda}$ represents a quaternion multiplication, $\mathbf{R}(\mathbf{q})$ the quaternion rotation, $m$ the quadrotor's mass, and $\mathbf{J}$ its inertia.

\subsection*{Quadrotor Inputs}
\begin{figure}[t]
\centering
\vspace{-5pt}
\includegraphics[width=\linewidth]{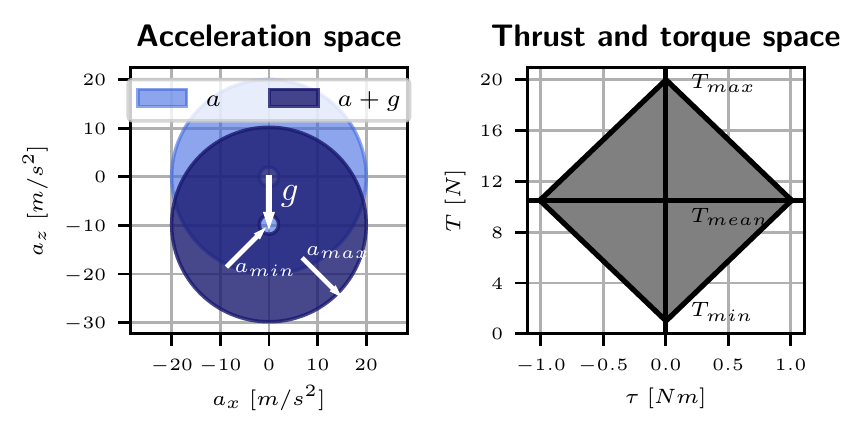}
\caption{\textbf{Quadrotor input space.} Acceleration (left) and thrust/torque-space (right) of a standard quadrotor configuration.
Note that the acceleration space is non-convex due to the minimum acceleration (the idle thrust at minimum motor speed) $a_{min} > 0$ being non-zero, and the thrust and torque limits are dependent on each other.}
\label{fig:inputs}
\end{figure}

The input space given by $\mathbf{T}$ and $\boldsymbol{\tau}$ is further decomposed into the single rotor thrusts $\mathbf{u} = [T_1, T_2, T_3, T_4]$. where $T_i$ is the thrust at rotor $i \in \{ 1, 2, 3, 4 \}$.
\begin{align}
\mathbf{T} &= \begin{bmatrix}0 \\ 0 \\ \sum T_i\end{bmatrix} &
&\text{and}&
\boldsymbol{\tau} &=
\begin{bmatrix}l/\sqrt{2} (T_1 + T_2 - T_3 - T_4) \\
l/\sqrt{2} (- T_1 + T_2 + T_3 - T_4) \\
c_\tau (T_1 - T_2 + T_3 - T_4)\end{bmatrix}
\end{align}
with the quadrotor's arm length $l$ and the rotor's torque constant $c_\tau$.
The quadrotor's actuators limit the applicable thrust for each rotor, effectively constraining $T_i$ as 
\begin{equation}
0 \leq T_{min} \leq T_i \leq T_{max} .
\label{eq:quad_limits}
\end{equation}
In Fig. \ref{fig:inputs} we visualize the acceleration space and the thrust torque space of a quadrotor in the $xz$-plane.
Note that the acceleration space in Fig. \ref{fig:inputs} is non-convex due to $T_{min} > 0$ for the depicted model parameters from the standard configuration of Tab. \ref{tab:quads}.
The torque space is visualized in Fig. \ref{fig:inputs}, where the coupling between the achievable thrust and torque is visible.

\subsection{Approximative Linear Aerodynamic Drag}
Finally, we extend the quadrotor's dynamics to include a linear drag model \cite{Faessler18ral}, to approximate the most dominant aerodynamic effects with diagonal matrix $\mathbf{D}$ by
\begin{equation}
\dot{\mathbf{v}} = \mathbf{g} + \frac{1}{m} \mathbf{R}(\mathbf{q}) \mathbf{T} - \mathbf{R}(\mathbf{q}) \mathbf{D} \mathbf{R}^\top(\mathbf{q}) \cdot \mathbf{v}
\end{equation}
where we approximate $\mathbf{D} = \text{diag}(d_x, d_y, d_z)$ in this work.
\section{Results}
\label{sec:exp_realworld}

\noindent
Video of the Results:\\
\url{https://youtu.be/ZPI8U1uSJUs} \\

\noindent
We chose drone racing as a demonstrator for our method because in racing the ultimate goal is to fully exploit the actuator potential to accomplish a task in minimal time.
In our experiment, we set up a human baseline on a 3D race track with 7 gates (Fig. \ref{fig:eyecatcher} \& \ref{fig:realworld_traj}) in a motion capture environment with two professional expert drone racing pilots.
We plan a time-optimal trajectory through the same race track and use an in-house developed drone platform and software stack to execute the trajectory in the same motion capture environment.
We generate the trajectory at a slightly lower thrust bound than what the platform can deliver, to maintain controllability under disturbances as mentioned in Section \ref{sec:pref_realworld} and discussed in Section \ref{sec:dis_applicability}.
Our results show that we can outperform the humans and consistently beat their best lap time.

\subsection{Experimental Drone Platform}
\label{sec:exp_real_drone}

\setcounter{figure}{1}
\begin{figure}[t]
    \centering
    \includegraphics[width=\linewidth,trim={90 220 80 160},clip]{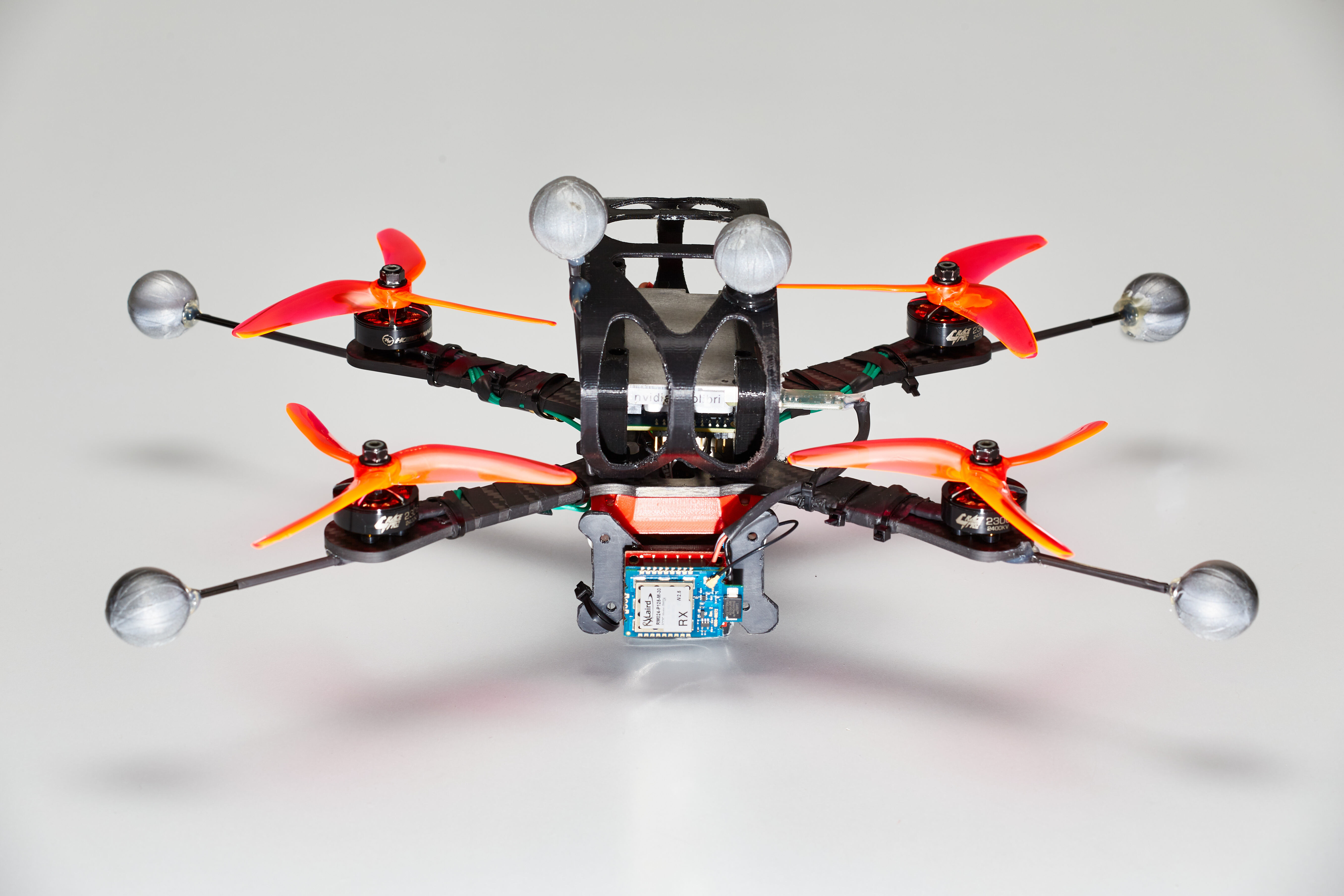}
    \caption{\textbf{The quadrotor vehicle.} The autonomous platform used for the real-world experiments with a theoretical thrust-to-weight ration of $\sim 4$ at $\SI{0.8}{\kilo\gram}$ weight, equipped with a Jetson TX2, a Laird communication module, off-the-shelf drone racing components, and infrared-reflective markers for motion capture.}
    \label{fig:drone}
\end{figure}

The experiments are flown with an in-house developed drone platform based on off-the-shelf drone-racing components such as a carbon-fiber frame, BLDC motors, $5"$ propellers, and a BetaFlight flight controller.
The quadrotor is equipped with an NVIDIA Jetson TX2 compute unit with WiFi and a Laird RM024 module for wireless low-latency communication.
The static maximum thrust of a single rotor was measured using a load cell and verified in-flight, marking the platform's real maximum limit at a thrust to weight ratio of roughly $\sim4$.
Other specifications can be taken from Tab. \ref{tab:quads} under the \textit{Race Quad} configuration.

For state estimation (pose, linear and angular velocities), we use a VICON system with 36 cameras.
For control, we deploy a Model Predictive Controller, similar to \cite{Falanga18iros} but based on the quadrotor dynamics from (\ref{eq:quad_dynamics}--\ref{eq:quad_limits}), with the state space $\mathbf{x} = [\mathbf{p}_{IB}, \mathbf{q}_{IB}, \mathbf{v}_{IB}, \boldsymbol{\omega}_{B}]^\top$ and input space $\mathbf{u} = [T_1, T_2, T_3, T_4]$.
The MPC operates over a horizon of $N_{MPC} = 20$ time steps of $\delta t = \SI{0.05}{\second}$, with a quadratic cost function, and also accounts for the single-rotor thrust constraints.
The implementation is done using the ACADO\cite{Houska2011ocam} toolkit and qpOASES\cite{Ferreau2014mpc} as solver.
We execute the MPC in a real-time iteration scheme \cite{Diehl2006springer} at a feedback rate of $\SI{100}{\hertz}$.
The low-level BetaFlight controller has access to high-frequency IMU measurements, which allows precise tracking of bodyrate and collective thrust commands.
These commands are extracted from the MPC controller and are guaranteed to stay within the platform capabilities due to the single-rotor thrust constraints.

\begin{table}[t]
    \centering
    \setlength{\tabcolsep}{3pt}
    \caption{\textbf{Quadrotor Configurations}}
    \label{tab:quads}
    \begin{tabular}{l|c|c|c}
        \hline
        Property & Race Quad & Airsim Quad & Standard Quad \\
        \hline
        $m$ $[\si{\kilo\gram}]$ & $0.8$ & $1.0$ & $1.0$ \\
        $l$ $[\si{\meter}]$ & $0.15$ & $0.23$ & $0.15$ \\
        $diag(J)$ $[\si{\gram\meter^2}]$ & $[1, 1, 1.7]$ & $[10, 10 ,20]$ & $[5, 5, 10]$ \\
        $[T_{min}, T_{max}]$ $[\si{\newton}]$ & $[0.0, 8.0]$ & $[0.0, 4.179]$ & $[0.25, 5.0]$ \\
        $c_\tau$ $[1]$ & $0.01$ & $0.0133$ & $0.01$ \\
        $\omega_{max}$ $[\si{\radian\per\second}]$ & $15$ & $10$ & $10$ \\
        $d_{drag}$ $[\si{\per\second}]$ & $0.4$ & $0.6$ & $-$ \\
        \hline
    \end{tabular}
    \vspace{-8pt}
\end{table}

\subsection{Human Expert Pilot Baseline}
Since human pilots so far outperformed autonomous vehicles, we establish a baseline by inviting two professional expert drone racing pilots, Michael Isler and Timothy Trowbridge, both of which compete in professional drone racing competitions.
A list of their participation and rankings can be found in Tab. \ref{tab:pilots} in the supplementary material.
We created a 3D racing track in a VICON motion capture environment spanning roughly $25 \times 30 \times 8 \si{\meter}$ and let the humans train on this track for hours.
We captured multiple races, each consisting of multiple laps, from which we evaluated the one with the overall best lap time, according to our timing strategy described in Section \ref{sec:realworld_timings}.
The quadrotor platform used for human flights has the same thrust-to-weight ratio as the autonomous platform.
This provides a fair baseline since (I) both the humans and the autonomous drone have the same limitations, (II) the humans are given enough training time to adapt to the track, as is the case at a real drone racing event, and (III) we compare against the race including the absolute best lap time from all runs.
Especially the latter point gives a substantial advantage to the human competitors, since they are typically not capable of reproducing the absolute best lap time reliably, and would therefore fall behind in any multi-lap evaluation.
The evaluated best human runs each contain 7 laps.

The human platform is also tracked in the VICON system and resembles the autonomous drone described in Section \ref{sec:exp_real_drone}, but drops the Jetson and Laird modules for a remote control receiver and a first-person-view camera system, with no weight difference.
To keep a fair baseline, the human platform is restricted to the same maximum thrust-to-weight limit as the autonomous drone.
The track, time-optimal reference (thrust-to-weight ratio $3.3$), and the human expert 1 trajectory are visualized in Fig. \ref{fig:realworld_traj}, with their speed and acceleration profile colorized.

\subsection{Trajectory Generation}
\begin{figure*}
    \centering
    \includegraphics[width=0.9\linewidth,trim={0 12 0 25},clip]{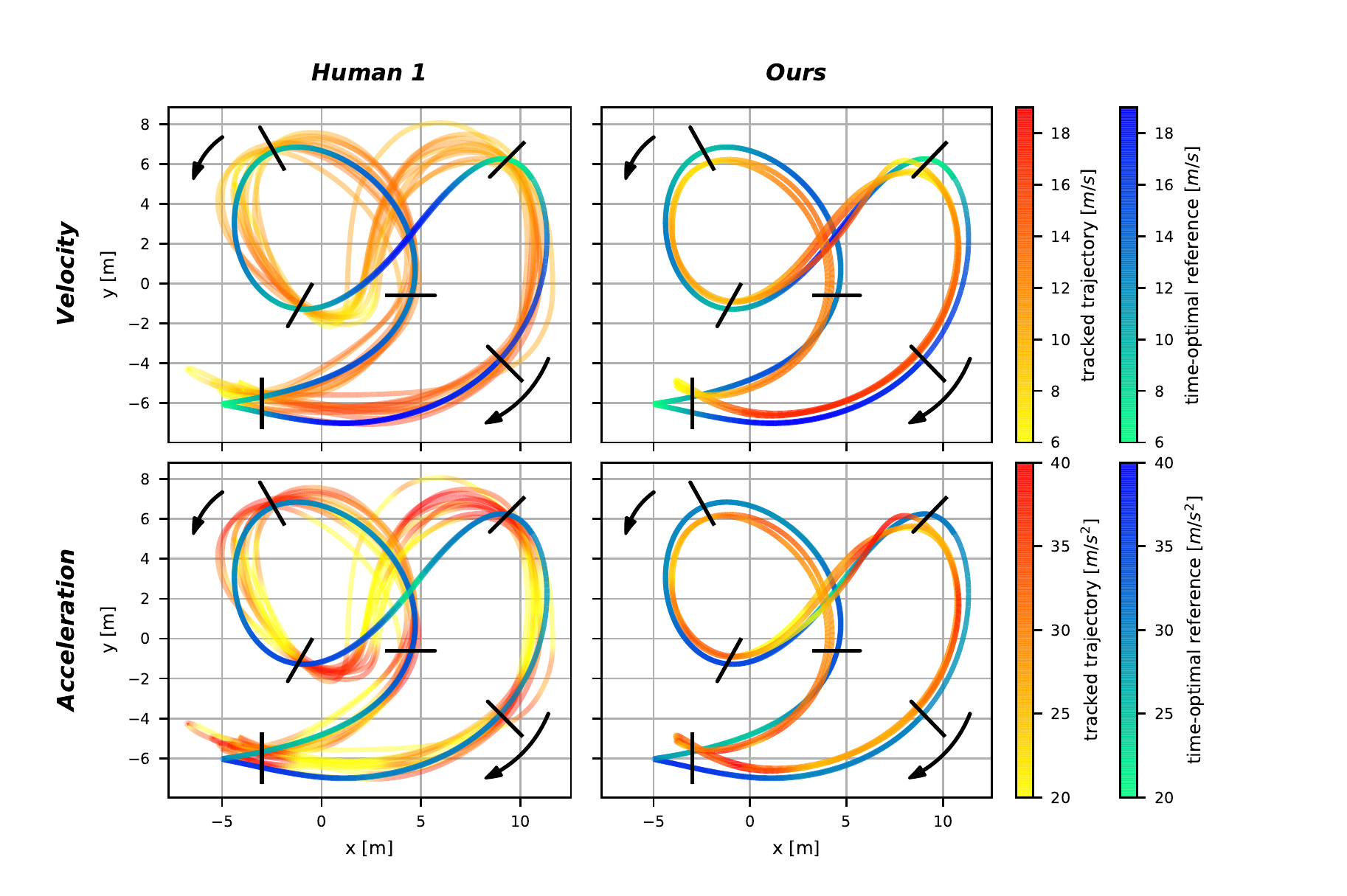}
    \caption{\textbf{Comparison against a human pilot.} The race track with seven gates visualized with the yellow-red real-world trajectories of the humans (left, all seven laps of the race including the overall best lap time) and the autonomous drone (right, two laps visualized), and the green-blue time-optimal reference trajectory with a thrust-to-weight ratio (TWR) of $3.3$.
    The trajectories are colored by their speed profile (top row) and acceleration profile (bottom row) to indicate the hotspots of highest velocity and acceleration along the track. The two black arrows indicate the direction of flight.
    The human pilots vary their acceleration substantially more than the autonomous drone and spend more time at sub-optimal acceleration (yellow coldspots on the lower left figure).
    Note that the gate in the lower-left corner consists of two gates stacked vertically.}
    \label{fig:realworld_traj}
\end{figure*}

To generate the time-optimal trajectories that will be executed by our platform, we set the gate positions of the track shown in Fig. \ref{fig:realworld_traj} as waypoint constraints.
We generate 2.5 laps to ensure that our experiments are not affected by start and end transient effects or large drops in battery voltage, and to ensure at least two full laps at maximum speed.
For optimal results, the laps are generated by concatenating the waypoints for a single lap multiple times and solving the full multi-lap problem at once.
Therefore, the optimization passes through 18 waypoints and can be solved on a normal desktop computer in $\sim\SI{40}{\minute}$.
We used $N=720$ nodes with a tolerance of $d_{tol}=\SI{0.3}{\meter}$.

Because a time-optimal trajectory exploits the full actuator potential, it is extremely aggressive from start to end.
To ensure safe execution on a real drone, we linearly ramp up the thrust limit for the trajectory generation from hover to the full thrust limit, guaranteeing a smooth start of the trajectory.
Since we plan over 2.5 laps and exclude the start and end from the timing, this has no notable impact on the reported timing results.
The same start and end exclusion are done for the human timings (Section \ref{sec:realworld_timings}).

Additionally, we plan the trajectory for a multitude of thrust-to-weight ratios reaching from $2.5$ with a lap time of $\SI{7.14}{\second}$ to a maximum of $3.6$ resulting in a lap time of $\SI{5.81}{\second}$.
As expected, the time shrinks with higher thrust-to-weight capabilities.
We evaluate two configurations with thrust-to-weight ratios of $3.15$ ($\SI{6.27}{\second}$) and $3.3$ ($\SI{6.10}{\second}$) in our real-world experiments, reported in the following section.
Both of these configurations are consistently faster than the human trajectory while staying within a safe margin of the quadrotors absolute limit (TWR: $\sim4$), which allows for robust control even under disturbances, noise, and model imperfections.

\subsection{Timing Analysis}~\\
\label{sec:realworld_timings}
First of all, our real-world experiments should provide a proof-of-concept that time-optimal trajectories planned below the drone's actual thrust limit are a feasible and viable solution.
We refer the reader to the accompanying Movie at \url{https://youtu.be/ZPI8U1uSJUs}.

\begin{figure}[t]
    \centering
    \includegraphics[width=\linewidth]{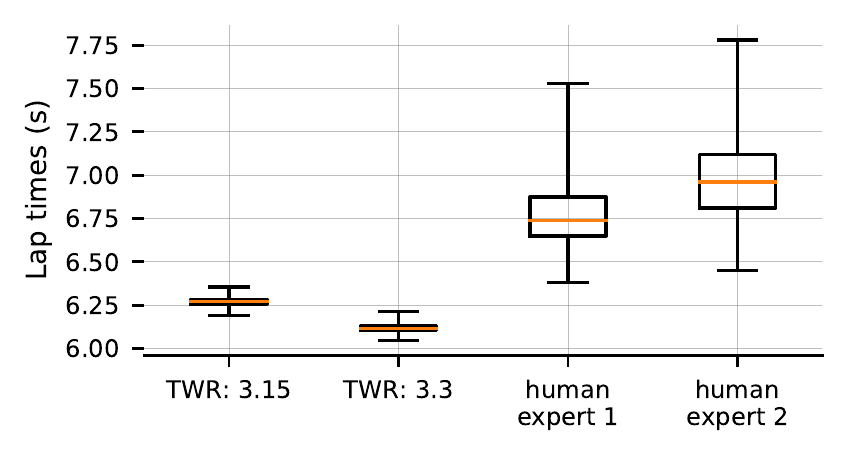}
    \caption{\textbf{Timing analysis.}
    Lap time box plot of our time-optimal trajectory executed in real-world for two configurations (two laps), and two human races (seven laps).
    Note that even with a $5\%$ reduced thrust-to-weight ratio of $3.15$, the proposed time-optimal trajectories are faster than the best human lap time, with substantially lower variance.
    }
    \label{fig:exp_laptimes}
\end{figure}

Second, we point out that it is possible to generate and execute trajectories that can outperform the human baseline.
For this we provide a statistical lap-time analysis in Fig. \ref{fig:exp_laptimes}, indicating the superior performance on both configurations.

To compute reliable lap times, we first define a full lap as each lap that is not affected by any start (take-off) or end (landing) segment.
We can then pick a set of timing points $\mathcal{S}_{pt} = \{\mathbf{p}_{t0}, \mathbf{p}_{t1}, \dots \}$ along the trajectory, and define the timing as the time needed to visit one such point twice, i.e. the time of a segment starting and ending at the same point.
This measure allows us to extract a statistically valid timing of a single closed lap, which does not depend on the location of timing start and stop.

Effectively we time the best laps of the human flights, and two full laps of both our race trajectories, each flown twice.
Fig. \ref{fig:exp_laptimes} shows the obtained timing results, where it is clearly visible that the autonomous drone outperforms the human pilots both in absolute time, but also consistency.
The latter is expected since once the trajectory is generated, it can be repeated multiple times without variation.

% \begin{table}[t]
%     \centering
%     \caption{\textbf{Timing Statistics}}
%     \label{tab:timing_stats}
%     \begin{tabular}{l|ccccc}
%         \hline
%         Timing & mean $[\si{\second}]$ & median $[\si{\second}]$ & min $[\si{\second}]$ & max $[\si{\second}]$ & std $[\si{\second}]$ \\
%         \hline
%         human expert 2 & $6.987$ & $6.960$ & $6.450$ & $7.780$ & $0.2859$ \\
%         \hline
%     \end{tabular}
% \end{table}

\begin{figure}[t]
    \centering
    \includegraphics[width=3.5in]{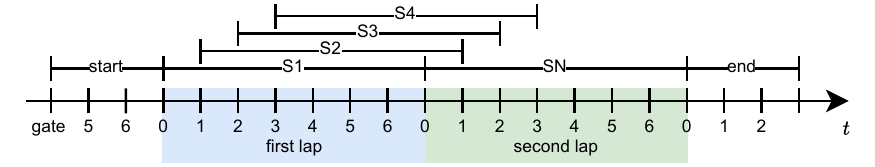}
    \vspace{6pt}
    \caption{\textbf{Visualization of the timing strategy.} A race includes two or more laps, over each of which we distribute 40 points at which we take timings of the resulting segments $S1, \dots, SN$. This allows for statistically meaningful timing extraction.}
    \label{fig:timing}
\end{figure}

\begin{table}[t]
    \centering
    \caption{\textbf{Timing Statistics}}
    \label{tab:timing_stats}
    \begin{tabular}{l|cccc}
        \hline
        \multirow{2}{*}{Timing} & \multirow{2}{*}{human 1} &\multirow{2}{*}{human 2} & ours & ours \\
        &&& TWR: $3.15$ & TWR: $3.3$ \\
        \hline
        mean $[\si{\second}]$ & $6.794$ & $6.987$ & $3.15$ & $6.120$ \\
        median $[\si{\second}]$ & $6.740$ & $6.960$ & $6.272$ & $6.116$ \\
        min $[\si{\second}]$ & $6.389$ & $6.450$ & $6.190$ & $6.048$ \\
        max $[\si{\second}]$ & $7.530$ & $7.780$ & $6.354$ & $6.215$ \\
        std $[\si{\second}]$ & $0.2556$ & $0.2859$ & $0.0280$ & $0.0278$ \\
        \hline
    \end{tabular}
\end{table}

%%%%%%%%%%%%%%%%%%%%%%%%%%%%%%%%%%%%%%%%%%%%%%%%%%%%%%%%%%%%%%%%%%%%%%%%%%%%%%%%

\section{Discussion}
\label{sec:discussion}

\subsection{Velocity and Acceleration Distribution}~\\
\label{sec:dis_velaccprofile}
Furthermore, we evaluate the velocity and acceleration distribution over the different human and time-optimal flights.
We depict the trajectories colored by their speed and acceleration profile in Fig. \ref{fig:realworld_traj} with the time-optimal reference with TWR: $3.3$.
Inspecting the hotspots of darker colors in the velocity plots (Fig. \ref{fig:realworld_traj}, top row) indicating higher speeds, we can see that the velocity distribution is rather similar for the autonomous time-optimal and the human trajectory.
However, comparing the acceleration of the human and the time-optimal trajectory (Fig. \ref{fig:realworld_traj}, bottom row), we observe that the acceleration of the human varies considerably more, and often dips to lower values than the time-optimal ones.
This corresponds to sections where the human pilots do not use the full actuation spectrum of the platform and lose substantial performance over the time-optimal trajectory.
This is especially visible in the right-most section of the track, where the flight path has relatively low curvature.
The time-optimal trajectory exploits the full acceleration capabilities, whereas the human notably reduces the acceleration between the right-most gates, leading to lower speeds in the following bottom section of the track.

Furthermore, we can identify the high-speed region in the sections of low curvature, where both human and autonomous platforms spend more acceleration in the velocity direction (accelerating and braking), than perpendicular to the velocity (direction change).
Although the platforms have equal TWR, and our time-optimal trajectory is planned with a substantial margin to the platform's TWR limit, it exceeds the human speed profile.

\subsection{Human Performance Comparison}

From our findings in Section \ref{sec:realworld_timings} and \ref{sec:dis_velaccprofile}, we conclude that human pilots are not far away from our time-optimal trajectory, since they reach similar but slower velocity distribution and lap times.
However, humans struggle to consistently exploit the full actuation spectrum of the vehicle, resulting in suboptimal performance compared to our approach.
One possible reason for this can be found in \cite{Pfeiffer2021ral}, where we analyzed eye gaze fixations of human pilots during racing and found that they fixate their eyes on upcoming gates well before passing the next gate, indicating that humans use a receding planning horizon, whereas our approach optimizes the full trajectory at once.

\subsection{Tracking Error Considerations}
We want to point out that while we achieved successful deployment on a real quadrotor system, we experience substantial tracking errors in doing so, as visible in Fig. \ref{fig:realworld_traj}.
While this work is not about improving the tracking performance, but should rather serve as a feasibility study given our method, we still feel responsible for pointing out the encountered difficulties.
A number of effects lead to this tracking error of $\sim\SI{0,7}{\meter}$ positional RMSE:

(I)
We did not account for any latency correction of the whole pipeline, including motion capture and pose filtering, data transmission to the drone, MPC execution time, and communication to the flight controller.

(II)
Our simple linear-drag aerodynamic model was verified in \cite{Faessler18ral} at speeds up to $\SI{5}{\meter\per\second}$.
However, in the vastly higher speed regimes we reach during our real-world experiments, the model seems to be inaccurate, especially in terms of drag at higher speeds and in body $z$ direction.
The effect of this is visible in Fig. \ref{fig:realworld_traj}, where there is a disturbance towards the inside of the curvature for most of the time.

(III)
The used BetaFlight controller is a reliable system for human drone pilots.
However, as such, it includes many filtering, feed-forward, and control strategies that are tuned for a consistent human \textit{flight-feeling}.
Unfortunately, this does not translate to accurate closed-loop control or desirable control-loop shaping and causes more harm than good when used with closed-loop high-level controllers such as our MPC.
Not only is it not possible to command the single rotor thrusts from the MPC to BetaFlight, but the provided bodyrate tracking also performed poorly.
This, however, could be solved with a custom flight controller and might be part of further studies.

\subsection{Convexity and Optimality}
\label{sec:dis_convexity}
While the problem of trajectory optimization quickly becomes non-convex when using complex and/or non-linear dynamic models, constraints, or even cost formulations (e.g. obstacle avoidance), it is often a valid approach to generate feasible (in terms of model dynamics) trajectories.
In fact, given a non-convex problem, solution schemes such as the used interior point method can only guarantee local optimality and global feasibility, but not global optimality.
Within our approach, we can summarize the following non-convex properties:
\begin{itemize}
    \item The quadrotor dynamics (\ref{eq:quad_dynamics}) are non-linear and therefore non-convex in the context of \eqref{eqn:ms_forwardeuler}.
    \item The acceleration space of a quadrotor (\ref{eq:quad_limits}) is non-convex given non-zero minimal thrust $T_{min} > 0.0$.
    \item Eqn. (\ref{eq:cpc}) is non-convex due to the norm on distance and the bilinearity of the progress change $\boldsymbol{\mu}$ and tolerance slack $\boldsymbol{\nu}$.
\end{itemize}
In our experiments in Sections \ref{sec:exp_nonconv} and \ref{sec:exp_localglobal}, we provoke such non-convex properties, and explain how the optimization can be supported with an advanced initialization scheme to start close to the global optimum.

This can be achieved by reducing the non-linear quadrotor model into a linear point-mass model with bounded 3D acceleration input $\mathbf{u} = \mathbf{a}$ where $\|\mathbf{a}\| \leq a_{max}$.
The linear model removes the prominent non-convex dynamics and allows us to find a solution from which the original problem with the quadrotor model can be initialized, both in terms of translational trajectory, and also with a non-continuous orientation guess based on the point-mass acceleration direction.
While this initial guess is not yet dynamically feasible for quadrotors due to the absent rotational dynamics, it serves as a valid initial guess in the convex region of the global-optimal translation space.
Finally, the rotational non-convexity can be resolved by solving multiple problems of different initializations, as demonstrated in Section \ref{sec:exp_nonconv}.

Last but not least, the reader should note that even in the case of a local (but not global) optimal solution, the vehicle dynamics are satisfied and the trajectory is dynamically feasible.

\subsection{Real-World Deployment}
\label{sec:dis_applicability}
There are three challenges when deploying our approach in real-world scenarios.

The first problem is posed by the nature of time-optimal trajectories themselves, as the true solution for a given platform is nearly always at the actuator constraints, and leaves no control authority.
This means that even the smallest disturbance could potentially have damaging consequences for the drone and render the remainder of the trajectory unreachable.
One has to define a margin lowering the actuator constraints used for the trajectory generation to add control authority and therefore robustness against disturbances.
However, this also leads to a slower solution, which is no longer the platform-specific time-optimal one.
In the context of a competition, this effectively becomes a risk-management problem with interesting connections to game theory.

Second, our method is computationally demanding, ranging from a few minutes ($<\SI{20}{\minute}$) for scenarios as in Sections \ref{sec:exp_p2p} and \ref{sec:exp_nonconv} towards an hour or more for larger scenarios such as Section \ref{sec:exp_realworld} with $\sim\SI{40}{\minute}$ and Section \ref{sec:exp_airsim} with $\sim\SI{65}{\minute}$ on a normal desktop computer.
However, this is highly implementation-dependent and could be vastly broken down to usable times, or precomputed for static race tracks and other non-dynamic environments.
Furthermore, our approach provides a method for finding the theoretical upper bound on performance, as a benchmark for other methods.

Third, we use a motion capture system to deploy our method.
However, in real-world scenarios, such high-performance off-board localization systems are barely ever available.
This necessitates the deployment using on-board state-estimation systems, such as visual-inertial odometry.
Unfortunately, these systems can suffer from high motion blur in such fast flight scenarios, and therefore need substantial further research and development to be of sufficient robustness for the purpose of time-optimal flight \cite{Foehn20rss,Moon19jirc}.
Despite those difficulties, we have demonstrated that our method \textit{can} be deployed and \textit{is} in fact substantially faster than human experts.
\section{Acknowledgments}

We thank the two professional human pilots, Michael Isler and Timothy Trowbridge, for their participation.
Without their incredible skills, establishing a fair human baseline would not have been possible.
\paragraph*{Ethics Approval}
The study protocol has been approved by the local ethics committee of the University of Zurich.
The participants gave their written informed consent before participating in the study and gave their written consent to be mentioned by name and listed with their competition participation record.
%
% \paragraph*{Author Contribution}
% P.F. developed and implemented the problem formulation using complementary progress constraints and designed and performed all simulation experiments.
% P.F. and A.R. performed all real-world experiments and wrote the manuscript.
% D.S. provided funding, contributed to the design and analysis of the experiments, and revised the manuscript.
%
% \paragraph*{Funding}
% This work was supported by the National Centre of Competence in Research (NCCR) Robotics through the Swiss National Science Foundation (SNSF) and the European Union’s Horizon 2020 Research and Innovation Program under grant agreement No. 871479 (AERIAL-CORE) and the European Research Council (ERC) under grant agreement No. 864042 (AGILEFLIGHT).
%
\paragraph*{Data and Materials}
All materials are described within this paper and completed by the dataset and code necessary to reproduce our results can be found in \cite{cpcDataset} at \\
\url{https://doi.org/10.5061/dryad.9kd51c5h7}.
A video of the flight performance can be found at \\
\url{https://youtu.be/ZPI8U1uSJUs}.

{\small
\balance
\bibliographystyle{unsrtnat}
\bibliography{references}
}

\clearpage
% \appendix
% \renewcommand{\thesection}{S\arabic{section}}
% \section*{\huge{Supplementary Materials}}
% \renewcommand{\thetable}{S\arabic{table}}
% \setcounter{table}{0}
% \renewcommand{\thefigure}{S\arabic{figure}}
% \setcounter{figure}{0}

\begin{appendices}
\section{Preface: Time-Optimal Quadrotor Trajectory}
\label{sec:preface}
\subsection{Point-Mass Bang-Bang}
Time-optimal trajectories encapsulate the best possible action to reach one or multiple targets in the lowest possible time.
We first investigate a point mass in $\mathbb{R}^3$ controlled by bounded acceleration $\mathbf{a} \in \mathbb{R}^3  ~|~ \|\mathbf{a}\| \leq a_{max}$, starting at rest position $\mathbf{p}_0$ and translating to rest position $\mathbf{p}_1$.
The time-optimal solution takes the form of a bang-bang trajectory over the time $t_{opt}$, accelerating with $a_{max}$ for $t_{opt}/2$, followed by decelerating with $a_{max}$ for $t_{opt}/2$.
A trajectory through multiple waypoints can be generated similarly by optimizing or sampling over the switching times and intermittent waypoint velocities.
For the general solution and applications, we refer to \cite{Pontryagin1962wiley} and \cite{LaValle2006book}.

\subsection{Bang-Bang Relation for Quadrotors}
\label{sec:pref_bangbang}
A quadrotor would exploit the same maximal acceleration by generating the maximum thrust.
However, due to the quadrotor's underactuation, it cannot instantaneously change the acceleration direction but needs to rotate by applying differential thrust over its rotors.
As a result, the linear and rotational acceleration, both controlled through the limited rotor thrusts, are coupled (visualized later in Fig. \ref{fig:inputs}).
Therefore, a time-optimal trajectory is the \textit{optimal tradeoff between rotational and linear acceleration}, where the thrusts only deviate from the maximum to adjust the rotational rates.
Indeed, our experiments confirm exactly this behavior (see e.g. Fig. \ref{fig:exp_straight}).

\subsection{Sub-Optimality of Polynomial Trajectories}
\label{sec:pref_poly}
\begin{figure}[b]
    \centering
    \vspace{-3pt}
    \includegraphics[width=\linewidth]{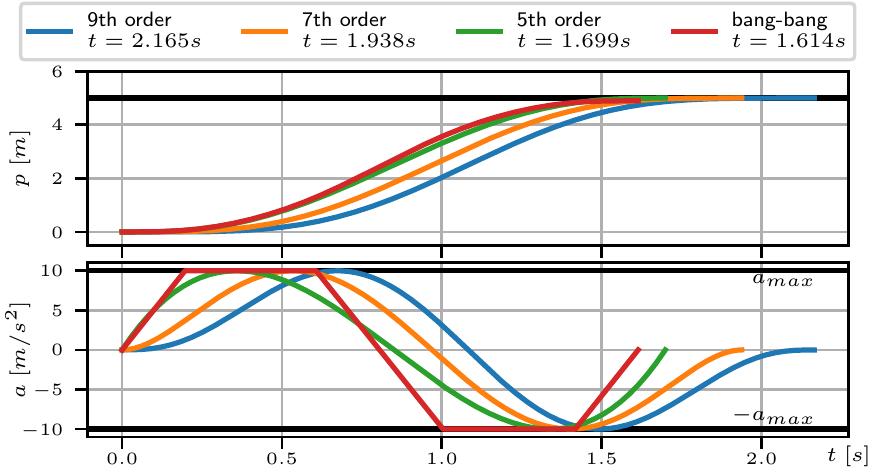}
    \caption{\textbf{Bang-bang and polynomial trajectories.}
    Multiple orders of polynomial trajectories and one bang-bang trajectory with limited slope.
    The polynomial trajectories only touch the input extrema in two points, while the bang-bang spends more time at the limit and achieves a lower overall time.}
    \label{fig:polyvsbang}
\end{figure}

The quadrotor is a differentially flat system \cite{Mellinger12ijrr} that can be described based on its four flat output states, position, and yaw.
This allows representing the evolution of the flat output states as smooth differentiable polynomials of the time $t$.
To generate such polynomials, one typically defines its boundary conditions at the start and end time, and minimizes for one of the derivatives, commonly the jerk or snap (3rd/4th derivative of position, as in \cite{Mueller13iros,Mellinger11icra}).
The intention behind such trajectories is to minimize and smooth the needed body torques and, therefore, single motor thrust differences, which are dependent on the snap.
Since the polynomials are very efficient to compute (especially \cite{Mueller13iros}), trajectories through many waypoints can be generated by concatenating segments of polynomials, and minimal-time solutions can be found by optimizing or sampling over the segment times and boundary conditions.
However, these polynomials are smooth by definition, which stands in direct conflict with maximizing the acceleration at all times while simultaneously adapting the rotational rate, as explained in the previous Section \ref{sec:pref_bangbang}.
In fact, due to the polynomial nature of the trajectories, the boundaries of the reachable input spaces can only be touched at one or multiple points, or constantly, but not at subsegments of the trajectory, as visualized in Fig. \ref{fig:polyvsbang}.

\subsection{Real-World Restrictions}
\label{sec:pref_realworld}
While the time-optimal solutions are at the boundary of the reachable input space and cannot be tracked robustly \cite{LaValle2006book}, they represent an upper bound on the performance for a given track, intended as a baseline for other algorithms.
However, we can lower the input bounds (w.r.t the real quadrotor actuator limit) for the purpose of trajectory generation, allowing for control authority and therefore restoring a margin for robustness, rendering the planned trajectories trackable in the real world.
As an application example, we present a demonstration of our method, where we track a time-optimal trajectory with a real quadrotor in a motion capture system and consistently outperform the human expert baseline.

\section{Simulation Experiments}
\label{sec:experiment_trajgen}
To demonstrate the capabilities and applicability of our method, we test it on a series of experiments.
We first evaluate a simple point-to-point scenario and compare to \cite{Hehn12ar,Loock13ecc} in Section \ref{sec:exp_p2p}.
Next, we investigate the time alignment for multiple waypoints in Section \ref{sec:exp_time}, followed by an experimental investigation of the convergence characteristics on short tracks in terms of initialization in Section \ref{sec:exp_initconv} and (non-) convexity in Section \ref{sec:exp_nonconv}.
Finally, we demonstrate applicability to longer tracks with $\geq 10$ waypoints (Section \ref{sec:exp_airsim}).
Note that this section will focus on the trajectory generation, while Section \ref{sec:exp_realworld} will demonstrate real-world deployment.

All evaluations are performed using CasADI \cite{Andersson18} with IPOPT \cite{Waechter06jmp} as solver backend.
We use multiple different quadrotor configurations, as listed in Tab. \ref{tab:quads}.
The first configuration represents a typical race quadrotor, the second one is parameterized after the MicroSoft AirSim \cite{Shah17fsr} SimpleFlight quadrotor, and the third standard configuration resembles the one used in \cite{Hehn12ar,Loock13ecc}.

\subsection{Initialization Setup}
If not stated differently, the optimization is initialized with identity orientation, zero bodyrates, $\SI{1}{\meter\per\second}$ velocity, linearly interpolated position between the waypoints, and hover thrusts.
The total time is set as the distance through all waypoints divided by the velocity guess.
The node of passing a waypoint (respectively where the progress variables $\boldsymbol{\lambda}$ switch to zero) is initialized as equally distributed, i.e. for waypoint $j$ the passing node is $k_j = N \cdot j / M$.
We define the total number of nodes $N$ based on the number of nodes per waypoint $N_w$ so that $N = M N_w$.
We typically chose roughly $N_w \in (50, 100)$ nodes per waypoint, to get a good linearization depending on the overall length, complexity, and time of the trajectory.
Note that high numbers of $N_w \gg 100$ help with convergence and achieved stability of the underlying optimization algorithm.

\begin{figure}[t]
    \centering
    \includegraphics[width=3.5in]{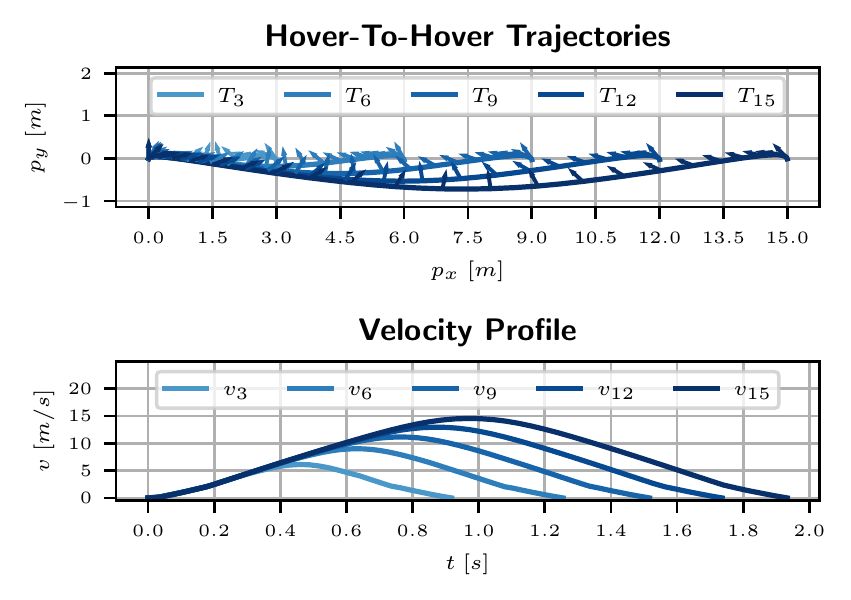}
    \caption{\textbf{Baseline comparison.} Time-optimal hover-to-hover trajectories between states spaced $p_x=[3, 6, 9, 12, 15]\si{\meter}$ apart as in \cite{Hehn12ar,Loock13ecc}.
    The top figure depicts the position on the $xz$-plane, while the bottom figure depicts the velocity profile.}
    \label{fig:exp_p2p}
\end{figure}

\subsection{Time-Optimal Hover-to-Hover Trajectories}
\label{sec:exp_p2p}
We first evaluate trajectory generation between two known position states in hover, one at the origin, and one at $p_x = [3, 6, 9, 12, 15]\si{\meter}$, as in \cite{Hehn12ar,Loock13ecc}.
Additionally to the problem setup explained in Section \ref{sec:method}, we add constraints to the end state to be in hover, i.e. $\mathbf{v}_N=\mathbf{0}$ and $\mathbf{q}=\begin{bmatrix}1 & 0 & 0 & 0\end{bmatrix}$.
We use $N=N_w=300$ nodes and a tolerance of $d_{tol}=10^{-3}$.
Different from \cite{Hehn12ar,Loock13ecc}, we compute the solution in full 3D space, which however does not matter for this experiment, since the optimal trajectory stays within the $y$-plane.
We defined the model properties so that it meets the maximal and minimal acceleration $[a_{min}, a_{max}] = [1, 20]\si{\meter\per\square\second}$ and maximal bodyrate $\omega_{max}=\SI{10}{\radian\per\second}$ as in \cite{Hehn12ar,Loock13ecc}, given by our standard quadrotor configuration (see Tab. \ref{tab:quads}).

\begin{table}[t]
    \centering
    % \scriptsize
    \begin{tabular}{lcccc}
        \hline
        $p_x$ vs Time & Ref. \cite{Hehn12ar} & Ref. \cite{Loock13ecc} & CPC-RT & CPC (ours) \\
        \hline
        $\SI{3}{\meter}$ & $\SI{0.898}{\second}$ & $\SI{0.890}{\second}$ & $\SI{0.891}{\second}$ & $\SI{0.918}{\second}$ \\
        $\SI{6}{\meter}$ & $\SI{1.231}{\second}$ & $\SI{1.223}{\second}$ & $\SI{1.227}{\second}$ & $\SI{1.255}{\second}$ \\
        $\SI{9}{\meter}$ & $\SI{1.488}{\second}$ & $\SI{1.478}{\second}$ & $\SI{1.484}{\second}$ & $\SI{1.517}{\second}$ \\
        $\SI{12}{\meter}$  & $\SI{1.705}{\second}$ & $\SI{1.694}{\second}$ & $\SI{1.702}{\second}$ & $\SI{1.736}{\second}$ \\
        $\SI{15}{\meter}$  & $\SI{1.895}{\second}$ & $\SI{1.885}{\second}$ & $\SI{1.894}{\second}$ & $\SI{1.933}{\second}$ \\
        \hline
    \end{tabular}
    \caption{Comparison of the resulting timings between our approach and \cite{Hehn12ar,Loock13ecc}. Note that our approach is $2.00\%$ slower than \cite{Hehn12ar} and $2.68\%$ slower than \cite{Loock13ecc}, because it accounts for rotation dynamics and single rotor limits.
    Our approach applied to the dynamics and limits of \cite{Hehn12ar,Loock13ecc}, closely reproduces their results (column \textit{CPC-RT}).}
    \label{tab:exp_p2p}
\end{table}

\begin{figure*}[t]
    \centering
    \includegraphics[width=\linewidth]{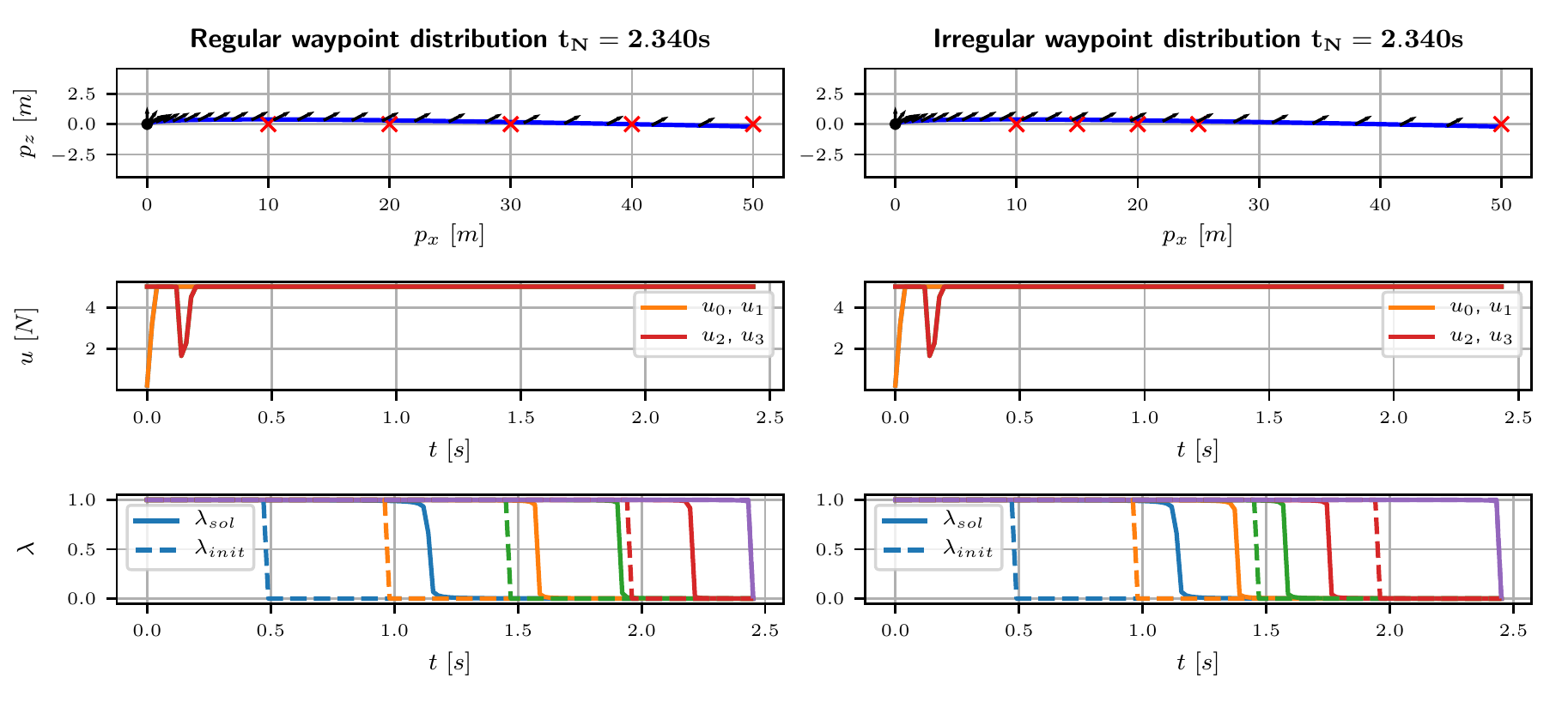}
    \caption{\textbf{Time allocation.}
    A trajectory along waypoints distributed on a line over $\SI{50}{\meter}$, flown in $\SI{2.430}{\second}$.
    On the left side, the waypoints are equally distributed over the total distance, while on the right side the first 4 out of 5 waypoints are located within the first half of the total distance.
    The bottom plot depicts the progress variables as initialized (dashed $--$) and as in the final solution (solid $-$).
    Note that both settings converge to the exact same solution, while the time of passing the waypoints was substantially adjusted from the initialization, and is different between the settings.
}
\label{fig:exp_straight}
\end{figure*}

The solutions are depicted in the $xy$-plane in Fig. \ref{fig:exp_p2p} and the timings are stated and compared to \cite{Hehn12ar,Loock13ecc} in Tab. \ref{tab:exp_p2p}.
In addition to our proposed single-rotor-thrust model \ref{eq:quad_dynamics} (reported as \textit{ours} in tab. \ref{tab:exp_p2p}), we also compute the timings for a simplified model with only collective thrust and bodyrate constraints, but no single-rotor-thrust modeling, as used in \cite{Hehn12ar,Loock13ecc} (reported as \textit{CPC-RT} in Tab/ \ref{tab:exp_p2p}).
Note that our approach with the simplified model produces results very similar to \cite{Hehn12ar,Loock13ecc}, but with the increased model-fidelity, our approach is $2.00\%$ slower than \cite{Hehn12ar} and $2.68\%$ slower than \cite{Loock13ecc}, due to the added rotational dynamics.
However, \cite{Hehn12ar,Loock13ecc} does not allow to compute trajectories through multiple waypoints and \cite{Spedicato18tcst} employs the same unrealistic bodyrate and thrust limits.
In contrast, our method accounts for both, realistic single rotor thrust limits and multiple waypoints, while simultaneously solving the time-allocation, as evaluated in the next section.

\subsection{Optimal Time Allocation on Multiple Waypoints}
\label{sec:exp_time}
In this experiment, we define a straight track between origin and $p_x = \SI{50}{\meter}$ trough multiple waypoints.
The goal is to show how our method can choose the optimal time at which a waypoint is passed.
Therefore, we test two different distributions of the waypoints $\mathbf{p}_{wj}$ over the straight track; specifically, we define a regular ($p_{x,reg} = [1, 20, 30, 40, 50]\si{\meter}$), and an irregular ($p_{x,ireg} = [10, 15, 20, 25, 50]\si{\meter}$) distribution.
We chose $N=125$ and a tolerance of $d_{tol}=\SI{0.4}{\meter}$, with the standard quadrotor (see Table \ref{tab:quads}).

As expected, both waypoint distributions converge to the same solution of $t_N = \SI{2.430}{\second}$, depicted in Fig. \ref{fig:exp_straight}, with equal state and input trajectories, despite the different waypoint distribution.
Since the waypoints are located at different intervals, we can observe a different distribution of the progress variables in Fig. \ref{fig:exp_straight}, while the trajectory time, dynamic states, and inputs stay the same.

\begin{figure*}[t]
    \centering
    \includegraphics[width=\linewidth]{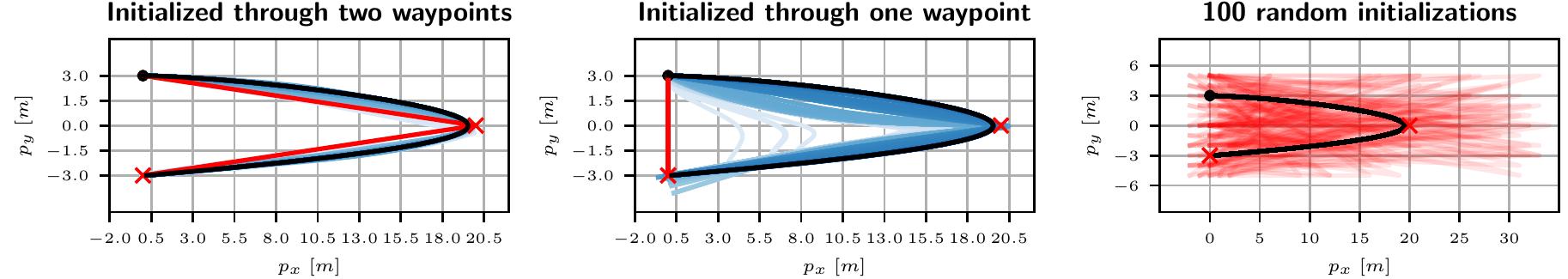}
    \caption{\textbf{Convergence quality.} Convergence in an open hairpin turn from different initializations in red ({\color{red} ---}) over the iterations in light blue (\textcolor[rgb]{0.45, 0.76, 0.98}{---}), to the final solution in dark blue(\textcolor[rgb]{0.0, 0.22, 0.66}{---}).
    The trajectory starts at the top left and passes through the two waypoints ({\color{red} $\times$}).
    While the good initialization in the left figure needs $216$ iterations, the poor guess in the middle figure needs $303$ iterations, but both converge to exactly the same solution with $t_N = \SI{3.617}{\second}$.
    The rightmost figure additionally shows $100$ random initializations, all converging to the same solution.}
    \label{fig:exp_hairpin}
\end{figure*}

\subsection{Initialization \& Convergence}
\label{sec:exp_initconv}
As a next step, the method is tested for convergence properties given different initializations.
For this, we again use the standard quadrotor configuration and discretize the problem into $N=160$ nodes with a tolerance of $d_{tol} = \SI{0.4}{\meter}$.
We use a track consisting of a so-called open hairpin, consisting of two waypoints as seen on the $xy$-plane in Fig.~\ref{fig:exp_hairpin}, starting on the top left and passing the waypoint to the far right and bottom left, where the endpoint is not in hover.
Multiple setups are tested, where the first one is initialized with the position interpolated between the waypoints (Fig.~\ref{fig:exp_hairpin}, left), the second one is initialized with a poor guess interpolated only from start to endpoint (Fig.~\ref{fig:exp_hairpin}, middle), and the third one includes 100 random initializations (Fig.~\ref{fig:exp_hairpin}, right).

The expected outcome is that all initialization setups should converge to the same solution.
Indeed we can observe this behavior in Fig.~\ref{fig:exp_hairpin}, where we depict the initial position guess in red and the convergence from light to dark blue.
The good initial guess in the leftmost Fig.~\ref{fig:exp_hairpin} results in $216$ iterations until convergence, while the poor guess in the middle Fig.~\ref{fig:exp_hairpin} needs $303$ iterations.
In the rightmost Fig.~\ref{fig:exp_hairpin} we perform 100 uniform random initializations on a $6 \times 6 \si{\meter}$ $xy$-plane around the start and end point, and a $30 \times 12 \si{meter}$ $xy$-plane around the mid point.
All initializations converge to the same solution.
Overall, this indicates that our method is not sensitive to initialization variations in the translation space, but profits from good guesses.
However, since the acceleration and orientation space of a quadrotor is not convex, the next two experiments elaborate on how to provoke and circumvent possible non-convexity issues in the rotation dynamics.

\subsection{Provoking Non-Convexity Issues}
\label{sec:exp_nonconv}
\begin{figure}[t]
    \centering
    \includegraphics[width=\linewidth]{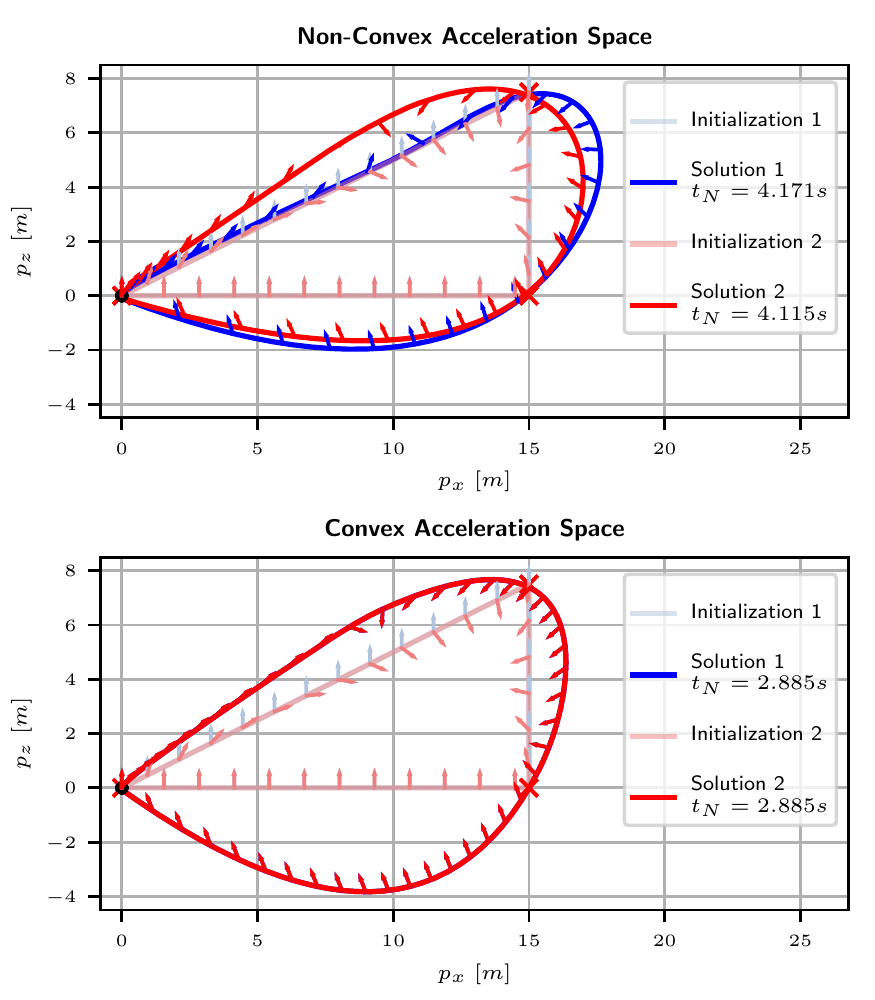}
    \caption{\textbf{Influence of convexity.} A vertical turn flown starting at the origin in hover and passing through both waypoints from top to bottom, and back to the origin.
    Two quadrotor configurations are used (standard in the top figure and race configuration in the bottom figure), with two different initialization setups each.
    The first setup is as described in Section \ref{sec:experiment_trajgen} with identity orientation, while the second setup uses a linearly interpolated orientation guess.
    The arrows indicate the thrust direction of the quadrotor.
    The standard quadrotor configuration converges to two different solutions in the top figure depending on the initialization due to its non-convex acceleration space with $T_{min} > \SI{0}{\newton}$, while the race quadrotor configuration converges to equal (and overlaying) solutions in the bottom figure, due to its convex acceleration space with $T_{min} = \SI{0}{\newton}$.
    }
    \label{fig:exp_vertpin}
\end{figure}

Since quadrotors can only produce thrust in body $z$-axis and the motors of most real-world systems cannot turn off, and therefore always produce a positive minimal thrust $T_{min} > 0$, the resulting acceleration space is non-convex.
We evaluate this on a vertical turn where we fly from hover at the origin through two waypoints directly above each other, back to the origin but not in hover.
The Track can be seen in Fig. \ref{fig:exp_vertpin}.

First, the standard quadrotor configuration is used, with $N=150$ nodes and a tolerance of $d_{tol}=\SI{0.1}{\meter}$, with the general initialization setup where the orientation is kept at identity.
A second setup uses a different initialization, where we interpolate the orientation around the $y$-axis between $\alpha_{init} = 0$, $\alpha_0 = \pi$ for the second waypoint and $\alpha_1 = \alpha_2 = 2 \pi$ for the remaining waypoints.
The solutions and associated initializations are depicted in the top Fig.~\ref{fig:exp_vertpin}, in which it is obvious that they do not converge to the same solution.
The second setup actually performs a flip which is slightly faster at $\SI{4.115}{\second}$ compared to the first setup at $\SI{4.171}{\second}$ ($1.3\%$ faster).
This is expected due to the non-convex properties of the problem.

However, a second set of experiments is performed with the same initialization setups but the race quadrotor configuration.
This configuration has a minimum thrust of $T_{min} = \SI{0}{\newton}$, which renders the achievable acceleration space convex.
Note that the full problem formulation still is non-convex, due to the non-linear dynamics and constraints.
Both setups for the race configuration are depicted in the bottom Fig.~\ref{fig:exp_vertpin}, and indeed, both initializations now converge to the same solution, which overlay each other and achieve the same timing at $\SI{2.885}{\second}$.

A simple solution would be to first solve the same problem using a linear and therefore convex point-mass model and using this to initialize the problem with the full quadrotor model.
We further elaborate on the convexity property in the discussion Section \ref{sec:dis_convexity}.

\subsection{Local-vs-Global Optimum}
\label{sec:exp_localglobal}
We follow up the previous non-convex example with another edge case where the non-convex dynamics can be used to provoke a local optimum, and we show how to resolve this special case.
The problem addressed is the pure-vertical translation from hover to hover, e.g. descending from an initial hover point at $\SI{5}{\meter}$ height, to the origin at $\SI{0}{\meter}$, parameterized with $N=100$ nodes, a tolerance of $d=\SI{0.1}{\meter}$, and using the race quad configuration.

\begin{figure}[b]
    \centering
    \includegraphics[width=\linewidth]{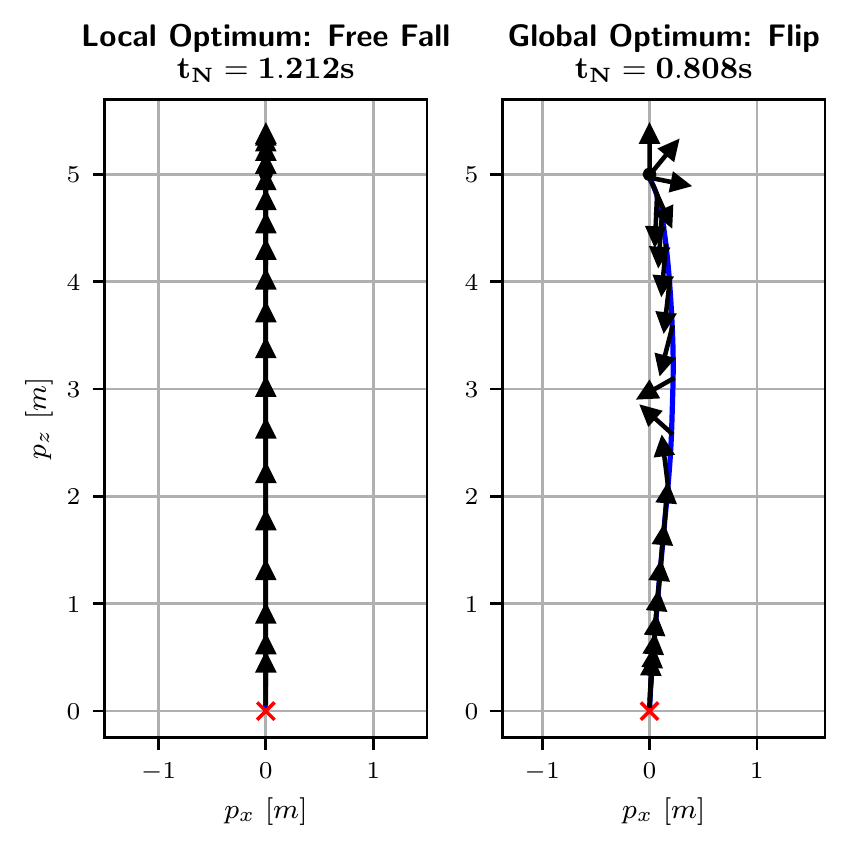}
    \caption{\textbf{Local and global optimum.} Two solutions to travel from a starting point at $\SI{5}{\meter}$ height to the origin at $\SI{0}{\meter}$.
    The left solution is initialized from an upright linear interpolated guess converging to a locally optimal free-fall descent.
    In contrast, the right solution is initialized from a bang-bang acceleration guess, reaching the global optimal solution of performing a flip.
    Note that the global optimal flip is notably faster ($\SI{0.808}{\second}$), than the locally optimal free fall ($\SI{1.212}{\second}$).}
    \label{fig:flipfall}
\end{figure}

When initialized with upright orientation and linear translation, the solution converges to a free-fall trajectory, which is a feasible local optimal solution, depicted in Fig. \ref{fig:flipfall}.
However, the optimal solution would exploit the available inputs in addition to the gravity acceleration, effectively performing a flip to accelerate downwards, before turning upright to decelerate into a hover state.
We can resolve this problem of local optimality by initializing the problem from a bang-bang point-mass guess.
This initial bang-bang guess exploits the full actuation space to accelerate for $t_{acc} = t_N / 2$ towards the end state, followed by symmetric deceleration phase of equal $t_{dec} = t_N / 2$, under the assumption of no gravity.
Essentially, this provides us with an initial guess for our orientation, which resolves the local optimum and converges to the global solution of performing a flip, visible in Fig. \ref{fig:flipfall}.
This initialization scheme can be applied for arbitrary trajectories to resolve most local optima.

\clearpage
\subsection{Microsoft AirSim, NeurIPS 2019 Qualification 1}
\label{sec:exp_airsim}
As an additional demonstration, we apply our algorithm on a track from the 2019 NeurIPS AirSim Drone racing Challenge \cite{Madaan20arxiv}, specifically on the Qualifier Tier 1 setup.
We choose a quadrotor with roughly the same properties, described as the MS configuration in Tab. \ref{tab:quads}.
The track is set up with the initial pose and 21 waypoints as defined in the environment provided in \cite{Madaan20arxiv}.
We use a discretization of $N = 3360$ nodes and a tolerance of $d_{tol} = \SI{0.1}{\meter}$.
The optimization of such a large state space took $\sim\SI{65}{\minute}$ on a normal desktop computer.

\begin{figure}[t]
    \centering
    \includegraphics[width=\linewidth]{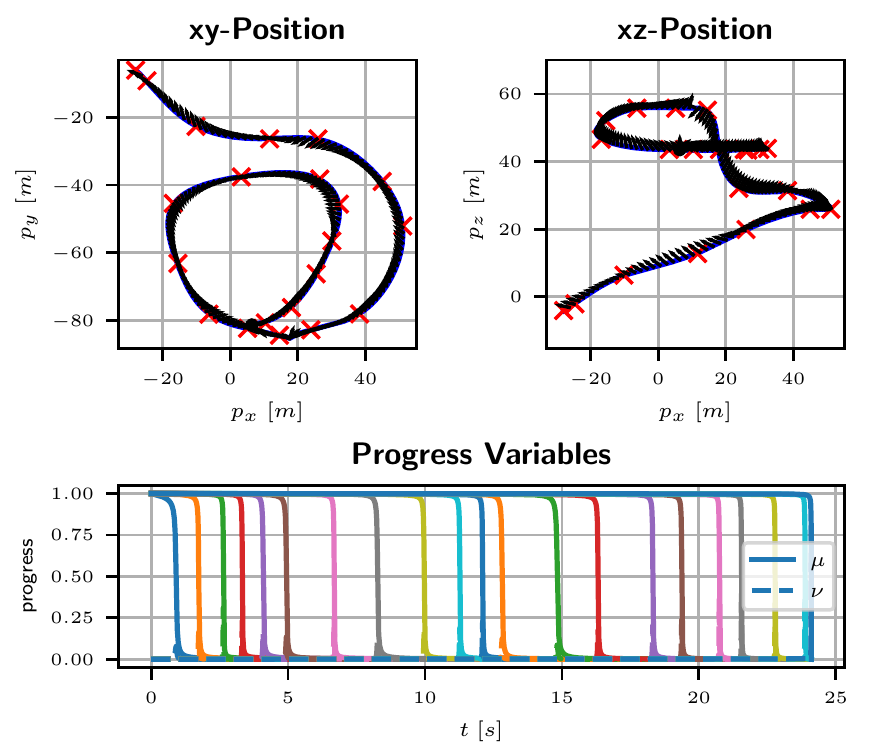}
    \caption{\textbf{Airsim qualification.} The NeurIPS Airsim Qualification 1 track, covered in $t_N = \SI{24.11}{\second}$, as opposed to the best team's $\SI{30.11}{\second}$.
    The top row depicts the trajectory in $xy$- and $xy$-plane, while the second row depicts the velocity, respectively, plotted over time.}
    \label{fig:exp_airsim_q1}
\end{figure}

The original work by \cite{Madaan20arxiv} provides a simple and conservative baseline performance of $t_{total} \approx \SI{110}{\second}$ under maximal velocity and acceleration of $v_{max} = \SI{30}{\meter\per\second}$ and $a_{max} = \SI{15}{\meter\per\second^2}$, respectively.
However, the best team achieved a time of $t_{total} = \SI{30.11}{\second}$ according to the evaluation page from \cite{Madaan20arxiv}.
Our method generates a trajectory that passes all waypoints at a mere $t_N = \SI{24.11}{\second}$, visualized in Fig. \ref{fig:exp_airsim_q1}.
Please note that this trajectory should only serve as a theoretical lower bound on the possibly achievable time given the model parameters.

\clearpage
\onecolumn
\vfill
\section{Human Pilot Rankings}
\begin{table}[h]
\small
\renewcommand{\arraystretch}{0.9}
\centering
\caption{Rankings}
\label{tab:pilots}
\begin{tabular}{lll}
\multicolumn{3}{c}{Michael Isler} \\
\toprule
Year & Event & Ranking \\
\midrule
\multirow{2}{*}{2020} &
Swiss Drone League Race Luzern & 2nd \\
& Highest Drone Race in the World -  St. Moritz & 1st \\
\midrule
\multirow{2}{*}{2019} &
Drone Champions League Laax & 3rd \\
& Swiss National Team FAI World Championship Shanghai & team \\
\midrule
\multirow{9}{*}{2018} &
Southern Germany Race & 3rd \\
& Drag Race Southern Germany & 3rd \\
& Swiss Online Freestyle Challenge & 2nd \\
& Kamikaze Race & 1st \\
& FAI Lausanne & 2nd \\
& Swiss Drone League Race St. Gallen & 1st \\
& Drone Champions League Rapperswil & 3rd \\
& Swiss National Team FAI World Championship Shenzen & team \\
& Drag Race at FAI World Championship Shenzen & 3rd \\
\midrule
\multirow{3}{*}{2017} &
Swiss Nationals & 2nd \\
& Swiss Indoor Masters & 2nd \\
& Drone Night & 1st \\
\bottomrule
&&\\
\multicolumn{3}{c}{Timothy Trowbridge} \\
\toprule
Year & Event & Ranking \\
\midrule
2020 & Drone Racing League & par \\
\midrule 
\multirow{7}{*}{2019}
& Drone Champions League Season & 2nd \\
& Drone Champions League Romania & 1st \\
& Drone Champions League Vaduz & 2nd \\
& Swiss Drone League Bern & 1st \\
& Swiss Drone League Luzern & 2nd \\
& Swiss Drone League Basel & 1st \\
& Drone Champions League Turin & 1st \\
\midrule
\multirow{8}{*}{2018}
& Drag Race at FAI World Championship Shenzen & 1st \\
& Drone Champions League Season & 1st \\
& Drone Champions League Zurich & 1st \\
& Drone Champions League Brussels & 1st \\
& Drone Champions League China & 1st \\
& XFly Great Wall of China & 1st \\
& Inter Copter Racing Cup & 2nd \\
& Thüringen Saison Opening & 2nd \\
\midrule
\multirow{6}{*}{2017}
& Moto Drone St. Gallen & 2nd \\
& Drone Champions League Brussels & 2nd \\
& Morph X Masters & 2nd \\
& Athens Drone GP & 1st \\
& Drone Champions League Paris & 2nd \\
& UpGreat Drone Race & 1st \\
\midrule
\multirow{2}{*}{2016}
& Danish Drone Nationals & 2nd \\
& Drone Night & 1st \\
\bottomrule
\end{tabular}
\\
\vspace{6pt}
Entries where the participants competed as part of a team are marked as "team",\\
while participation with undisclosed results are marked as "par".
\end{table}
\vfill
\end{appendices}

\end{document}